%% file: aaai25.tex
\documentclass[letterpaper]{article} 
\usepackage{aaai25}
\usepackage{times}  
\usepackage{helvet}  
\usepackage{courier}  
\usepackage[hyphens]{url}  
\usepackage{graphicx} 
\usepackage{natbib}  
\usepackage{caption} 
\usepackage{algorithm}
\usepackage{algorithmic}
\usepackage{multirow}
\usepackage{makecell}
\usepackage{subfig}
\usepackage{enumitem}
\usepackage{amsthm,amsmath,amssymb}
\usepackage{mathrsfs}
\usepackage{tikz}
\newcommand*\circled[1]{\tikz[baseline=(char.base)]{
            \node[shape=circle,draw,inner sep=2pt] (char) {#1};}}
\usepackage{newfloat}
\usepackage{listings}
\DeclareCaptionStyle{ruled}{labelfont=normalfont,labelsep=colon,strut=off} 
\floatstyle{ruled}
\newfloat{listing}{tb}{lst}{}
\floatname{listing}{Listing}
\title{MPQ-DM: Mixed Precision Quantization for Extremely Low Bit Diffusion Models}

\author {
    Weilun Feng\textsuperscript{\rm 1, \rm 2}\equalcontrib, 
    Haotong Qin\textsuperscript{\rm 3}\equalcontrib, 
    Chuanguang Yang\textsuperscript{\rm 1}\thanks{Corresponding authors.}, 
    Zhulin An\textsuperscript{\rm 1}\footnotemark[2], 
    Libo Huang\textsuperscript{\rm 1}, 
    Boyu Diao\textsuperscript{\rm 1}, \\
    Fei Wang\textsuperscript{\rm 1}, 
    Renshuai Tao\textsuperscript{\rm 4}, 
    Yongjun Xu\textsuperscript{\rm 1}, 
    Michele Magno\textsuperscript{\rm 3}
}
\affiliations {
    \textsuperscript{\rm 1}Institute of Computing Technology, Chinese Academy of Sciences \quad
    \textsuperscript{\rm 2}University of Chinese Academy of Sciences \\
    \textsuperscript{\rm 3}ETH Zurich \quad
    \textsuperscript{\rm 4}Beijing Jiaotong University\\
    \{fengweilun24s, yangchuanguang, anzhulin, diaoboyu2012, wangfei, xyj\}@ict.ac.cn, \\
    \{haotong.qin,michele.magno\}@pbl.ee.ethz.ch, www.huanglibo@gmail.com, rstao@bjtu.edu.cn
}

\usepackage{bibentry}

\begin{document}

\maketitle


\input{sec/0_abstract}

\input{sec/1_introduction}

\input{sec/2_related_work}

\input{sec/4_method}

\input{sec/5_experiment}

\input{sec/6_conclusion}

%

\section{Acknowledgements}
This work is partially supported by the National Natural Science Foundation of China (No.62476264 and No.62406312), China National Postdoctoral Program for Innovative Talents (No.BX20240385) funded by China Postdoctoral Science Foundation, Beijing Natural Science Foundation (No.4244098), Science Foundation of the Chinese Academy of Sciences, Swiss National Science Foundation (SNSF) project 200021E\_219943 Neuromorphic Attention Models for Event Data (NAMED), Baidu Scholarship, and Beijing Natural Science Foundation (L242021).

\bibliography{aaai25}

\newpage
\input{sec/supp}

\end{document}

%% file: sec/0_abstract.tex
\begin{abstract}
Diffusion models have received wide attention in generation tasks. However, the expensive computation cost prevents the application of diffusion models in resource-constrained scenarios. Quantization emerges as a practical solution that significantly saves storage and computation by reducing the bit-width of parameters. 
However, the existing quantization methods for diffusion models still cause severe degradation in performance, especially under extremely low bit-widths (2-4 bit). The primary decrease in performance comes from the significant discretization of activation values at low bit quantization. Too few activation candidates are unfriendly for outlier significant weight channel quantization, and the discretized features prevent stable learning over different time steps of the diffusion model. This paper presents \textbf{MPQ-DM}, a Mixed-Precision Quantization method for Diffusion Models.  
The proposed MPQ-DM mainly relies on two techniques:
(1) To mitigate the quantization error caused by outlier severe weight channels, we propose an \textit{Outlier-Driven Mixed Quantization} (OMQ) technique that uses $Kurtosis$ to quantify outlier salient channels and apply optimized intra-layer mixed-precision bit-width allocation to recover accuracy performance within target efficiency.
(2) To robustly learn representations crossing time steps, we construct a \textit{Time-Smoothed Relation Distillation} (TRD) scheme between the quantized diffusion model and its full-precision counterpart, transferring discrete and continuous latent to a unified relation space to reduce the representation inconsistency. 
Comprehensive experiments demonstrate that MPQ-DM achieves significant accuracy gains under extremely low bit-widths compared with SOTA quantization methods. MPQ-DM achieves a 58\% FID decrease under W2A4 setting compared with baseline, while all other methods even collapse.  
\begin{links}
\link{Code}{https://github.com/cantbebetter2/MPQ-DM}
\end{links}
\end{abstract}

%% file: sec/1_introduction.tex
\section{Introduction}
Diffusion models (DMs)~\cite{ho2020ddpm, dhariwal2021diffusionbeatgan} have demonstrated remarkable capabilities in generation tasks~\cite{rombach2022ldm, song2020score, mei2023vidm, yang2024diffusion, liu2023text}. 
However, the iterative denoising process and the massive of parameters in order to cope with the high-resolution demand seriously hinder the wide deployment of the diffusion model in edge devices with limited computing resources \cite{10.1093/comjnl/bxae110, DAI2024104811}.

\begin{figure}
\begin{minipage}{\linewidth}
    \centering
    \subfloat[][Baseline]{
        \includegraphics[width=0.5\linewidth]{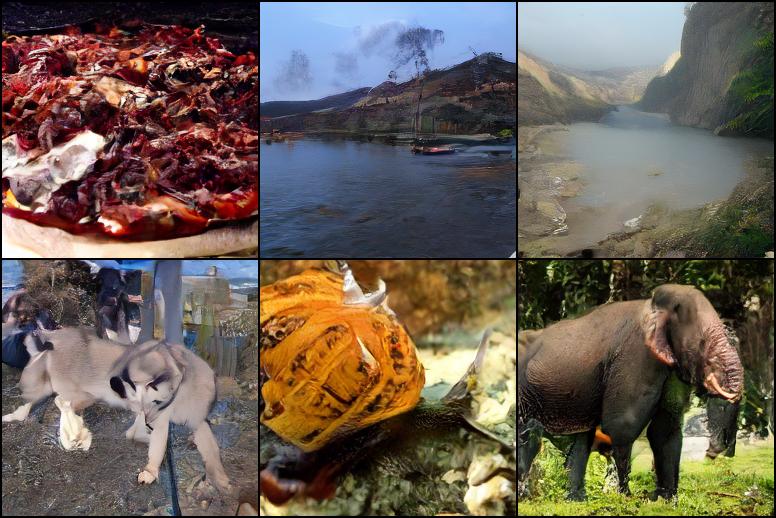}
        \label{}
    }
    \subfloat[][MPQ-DM]{
        \includegraphics[width=0.5\linewidth]{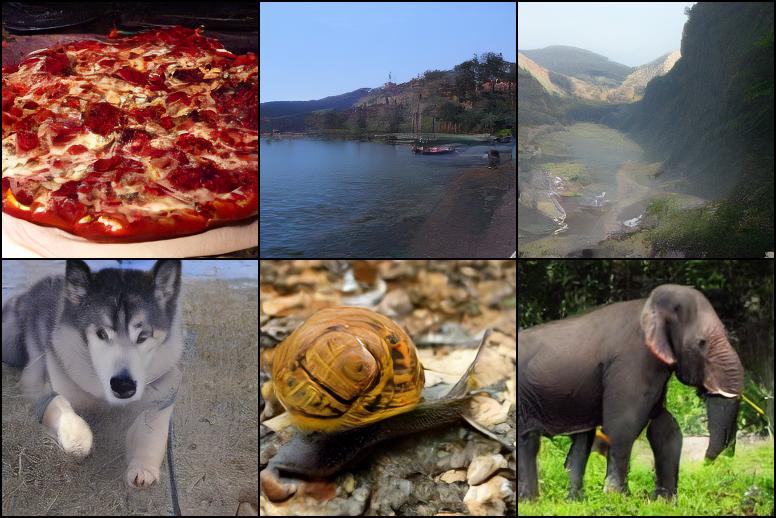}
        \label{}
    }
    \caption{Visualization of samples generated by baseline and MPQ-DM under W2A6 bit-width.}
    \label{fig:mse_with_kl}
\end{minipage}
\end{figure}

\begin{figure*}[!ht]
    \centering
    \includegraphics[width=\textwidth]{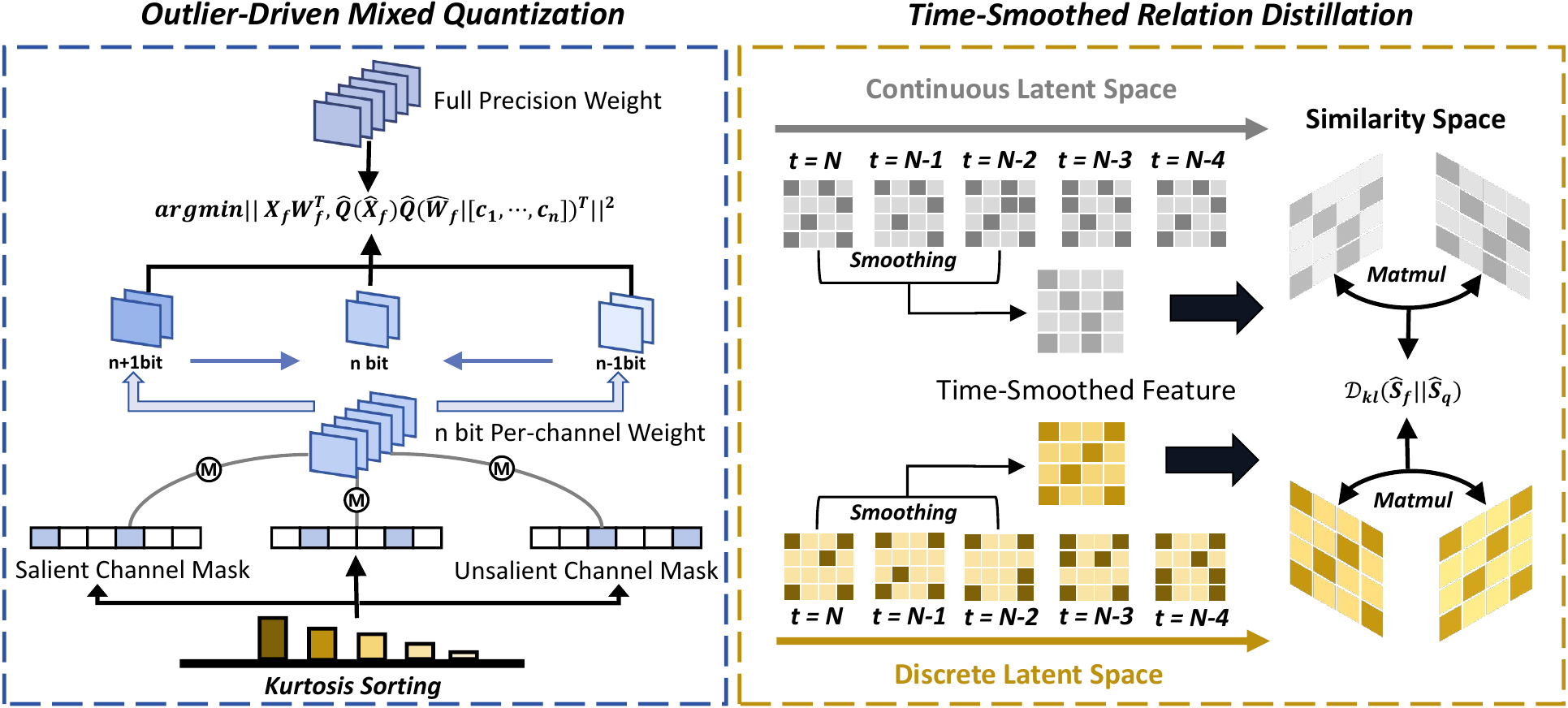}
    \caption{Overview of proposed MPQ-DM, consisting of Outlier-Driven Mixed Quantization to apply intra-layer mixed quantization and Time-Smoothed Relation Distillation to improve optimization robustness. \circled{M} stands for mask operation.}
    \label{fig:overview}
\end{figure*}
As an effective model compression approach that reduces the floating-point parameters to low-bit integers, model quantization can simultaneously reduce the model size and improve the inference speed \cite{gholami2022quantizationsurvey}. This technique has been widely used in CNNs \cite{pilipovic2018cnnquantsurvey, ding2024reg, chen2024scp} and Transformers \cite{chitty2023transformerquantsurvey}. 
Model quantization is mainly divided into post-training quantization (PTQ) \cite{hubara2020ptqlayer, wang2020towards, wei2022qdrop, liu2023pdquant} and quantization-aware training (QAT) \cite{esser2019lsq, jacob2018quantizationandtrain, krishnamoorthi1806quantizingwhitepaper}. 
QAT often requires a full amount of raw data to fine-tune model weights and quantization parameters. Therefore, it requires a fine-tuning time equivalent to original training but performs well on extremely low bit-width (2-4 bit) or even binarization \cite{qin2020binary, zheng2024binarydm}. PTQ only requires a small amount of calibration data to fine-tune the quantization parameters. Thus, a shorter calibration time is required, but model performance cannot be guaranteed. To solve this problem, quantization-aware low-rank fine-tuning (QLORA-FT) scheme \cite{he2023efficientdm} proposes to add a LoRA \cite{hu2021lora} module to perform a low-rank update on model quantized weights to enhance the performance of quantized diffusion models, to achieve a calibration time under PTQ-level but accurate quantization performance.

Despite exploring the QLORA-FT scheme in diffusion models, it still experienced significant performance degradation in extremely low bit quantization. The performance degradation mainly comes from the high discretization of activation values. We analyzed the challenges from two different aspects. 
From the representation perspective, we found significant outliers in some weight channels of the diffusion model, and the presence of outliers leads to a large number of outliers being occupied by stages or a small number of outliers in the bit width. Highly discretized activation values are extremely quantization-unfriendly for channels with severe outliers. This results in severe loss of weight information expression after quantization. Existing unified bit width quantization \cite{he2023efficientdm, wang2024quest} or layer-wise mixed precision quantization \cite{dong2019hawq, he2024ptqd, dong2020hawqv2} cannot solve outliers in target weight channel within layers. 
From the optimization perspective, highly discretized intermediate representation of quantization model resulting in not robust expression of features \cite{martinez2020trainingbinary}. The multi-step continuous denoising of the diffusion model leads to the accumulation of such errors. Also, the difference in latent space between the discretized features and the fully precision features may lead to negative optimization if we directly align two representations.

To address the aforementioned issues, we propose Mixed Precision Quantization for extremely low bit Diffusion Models (MPQ-DM) consisting of Outlier Driven Mixed Quantization (OMD) and Time Smoothed Relation Distillation (TRD). The overview of MPQ-DM is in Fig. \ref{fig:overview}. 
For channel-wise weight quantization, the weight quantization process of different channels does not affect each other. We use an outlier-aware scale to numerically mitigate the outlier degree.Then we use $Kurtosis$ to quantify the presence of outliers and assign higher quantization bits to channels with salient outliers, further ensuring its quantization performance. 
For model optimization, we select multiple consecutive intermediate representations as smoothed distillation targets to alleviate the problem of abnormal activation values in model optimization. To address the issue of numerical alignment mismatch between discrete and continuous latent spaces, we transfer the numerical alignment of the two latent spaces to a unified similarity space for knowledge distillation, ensuring consistency in feature expression. The contributions of our work are summarized as:
\begin{itemize}
    \item We identified salient outlier phenomena in different model weight channels as major bottlenecks for extremely low bit quantization. We propose Outlier-Driven Mixed Quantization, which utilizes smooth factor to alleviate outlier phenomenon numerically and dynamically allocates quantization bits of different channels within target bit-width.
    \item We identified extremely discretized features under extremely low bit quantization suffering from numerical unrobustness. We propose Time-Smoothed Relation Distillation to utilize the features of $N$ time steps as a smoothed distillation objective and transfer two latent spaces with large numerical differences to a unified similarity space for relation distillation.
    \item We push the limit of efficient diffusion quantization to extremely low bit-widths (2-4 bit). Extensive experiments on generation benchmarks demonstrate that our MPQ-DM surpasses both the baseline and current SOTA PTQ-based diffusion quantization methods by a significant margin.
\end{itemize}

%% file: sec/2_related_work.tex
\section{Related Work}
\subsection{Diffusion Model}
%
Diffusion models \cite{ho2020ddpm, rombach2022ldm} perform a forward sampling process by gradually adding noise to the data distribution $\mathbf{x}_0 \sim q(x)$. In DDPM, the forward noise addition process of the diffusion model is a Markov chain, taking the form:
\begin{equation}
\begin{gathered}
    q(\mathbf{x}_{1:T}|\mathbf{x}_0) = \prod \limits_{t=1}^T q(\mathbf{x}_t|\mathbf{x}_{t-1}), \\
    q(\mathbf{x}_t|\mathbf{x}_{t-1}) = \mathcal{N}(\mathbf{x}_t; \sqrt{\alpha_t}\mathbf{x}_{t-1}, \beta_t\mathbf{I}),
\end{gathered}
\end{equation}
where $\alpha_t=1-\beta_t$, $\beta_t$ is time-related schedule. Diffusion models generate high-quality images by applying a denoising process to randomly sampled Gaussian noise $\mathbf{x}_T \sim \mathcal{N}(\mathbf{0}, \mathbf{I})$, taking the form:
\begin{equation}
    p_{\theta}(\mathbf{x}_{t-1}|\mathbf{x}_t) = \mathcal{N}(\mathbf{x}_{t-1};\hat{\mu}_{\theta, t}(\mathbf{x_t}), \hat{\beta}_t\mathbf{I}),
\end{equation}
where $\hat{\mu}_{\theta, t}$ and $\hat{\beta}_t$ are outputed by the diffusion model. 

\subsection{Diffusion Quantization}
Post-training quantization (PTQ) and Quantization-aware
training (QAT) are two main approaches for model quantization. 
The commonly used QAT methods like LSQ  \cite{esser2019lsq} and methods for diffusion models Q-dm \cite{li2024qdm} and Binarydm \cite{zheng2024binarydm} ensure the model performance at extremely low bit-width or even binary quantization, but they require a lot of extra training time compared with PTQ methods, resulting larger training burden.
PTQ methods for diffusion model PTQ4DM \cite{shang2023ptq4dm} and Q-Diffusion \cite{li2023qdiffusion} have made initial exploration. The following works PTQ-D \cite{he2024ptqd}, TFMQ-DM \cite{huang2024tfmq}, APQ-DM \cite{wang2024apqdm} and QuEST \cite{wang2024quest} have made improvements in the direction of quantization error, temporal feature, calibration data, and calibration module. The performance of diffusion model after quantization is further improved. However, the performance of PTQ-based methods suffers from severe degradation at extremely low bit-width. 
To combine the advantages of QAT and reduce the required training time, EfficientDM \cite{he2023efficientdm} uses LoRA \cite{hu2021lora} method to fine-tune the quantized diffusion model.
However, neither of these efficient quantization methods can guarantee the performance of the diffusion model under low bit. Therefore, this paper focuses on maximizing the extremely low bit quantization diffusion models performance.

%% file: sec/4_method.tex
\section{Method}

\subsection{Model Quantization}
Model quantization maps model weights and activations to low bit integer values to reduce memory footprint and accelerate the inference. For a floating vector $\mathbf{x}_f$, the quantization process can be formulated as
\begin{equation}
\begin{gathered}
    \hat{\mathbf{x}}_q = Q(\mathbf{x}_f, s, z) = clip(\lfloor \frac{\mathbf{x}_f}{s} \rceil + z, 0, 2^N-1), \\
    s = \frac{u-l}{2^N-1}, z = - \lfloor \frac{l}{s} \rceil,
\end{gathered}
\end{equation}
where $\hat{\mathbf{x}}_q$ indicates quantized vector in integer, $\lfloor \cdot \rceil$ is round fuction and $clip(\cdot)$ is function that clamps values into the range of $[0, 2^N-1]$, $s$ is a scale factor and $z$ is a quantization zero point. $l$ and $u$ are the lower and upper bounds of quantization thresholds respectively. They are determined by $\mathbf{x}_f$ and the target bit-width.
Reversely, in order to restore the low bit integer quantization vector $\hat{\mathbf{x}}_q$ to the full precision representation, the dequantization process is formulated as
\begin{equation}
    \hat{\mathbf{x}}_f = \hat{Q}(\mathbf{x}_f) = (\hat{\mathbf{x}}_q - z)  s,
\end{equation}
where $\hat{\mathbf{x}}_f$ is the dequantized vector used for forward process.

\subsection{Outlier-Driven Mixed Quantization}
\begin{figure}
    \centering
    \subfloat[][Layer weight distribution.]{
        \includegraphics[width=0.5\linewidth]{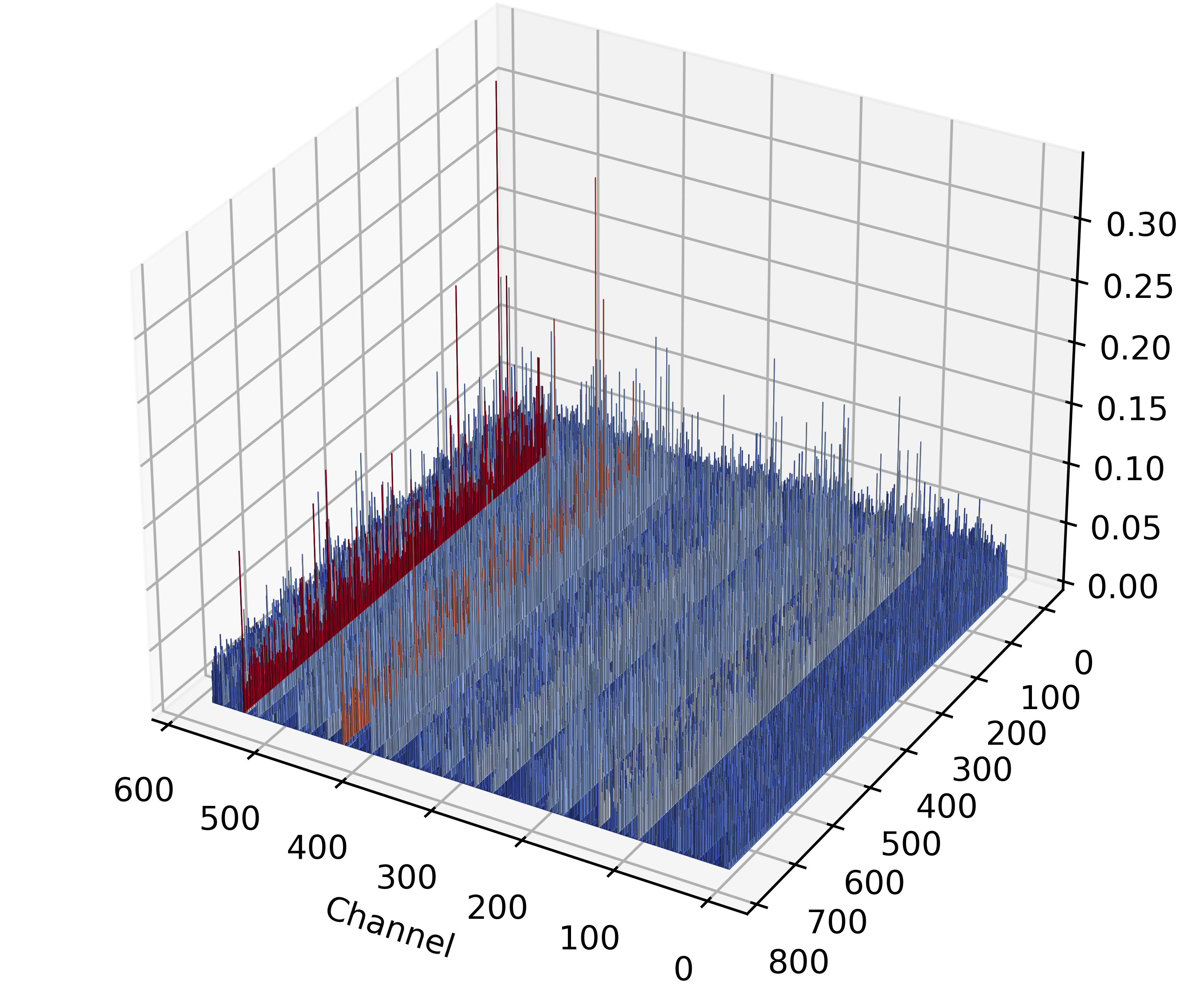}
        \label{fig:layer_weight}
    }
    \subfloat[][Outlier salient channel.]{
        \includegraphics[width=0.5\linewidth]{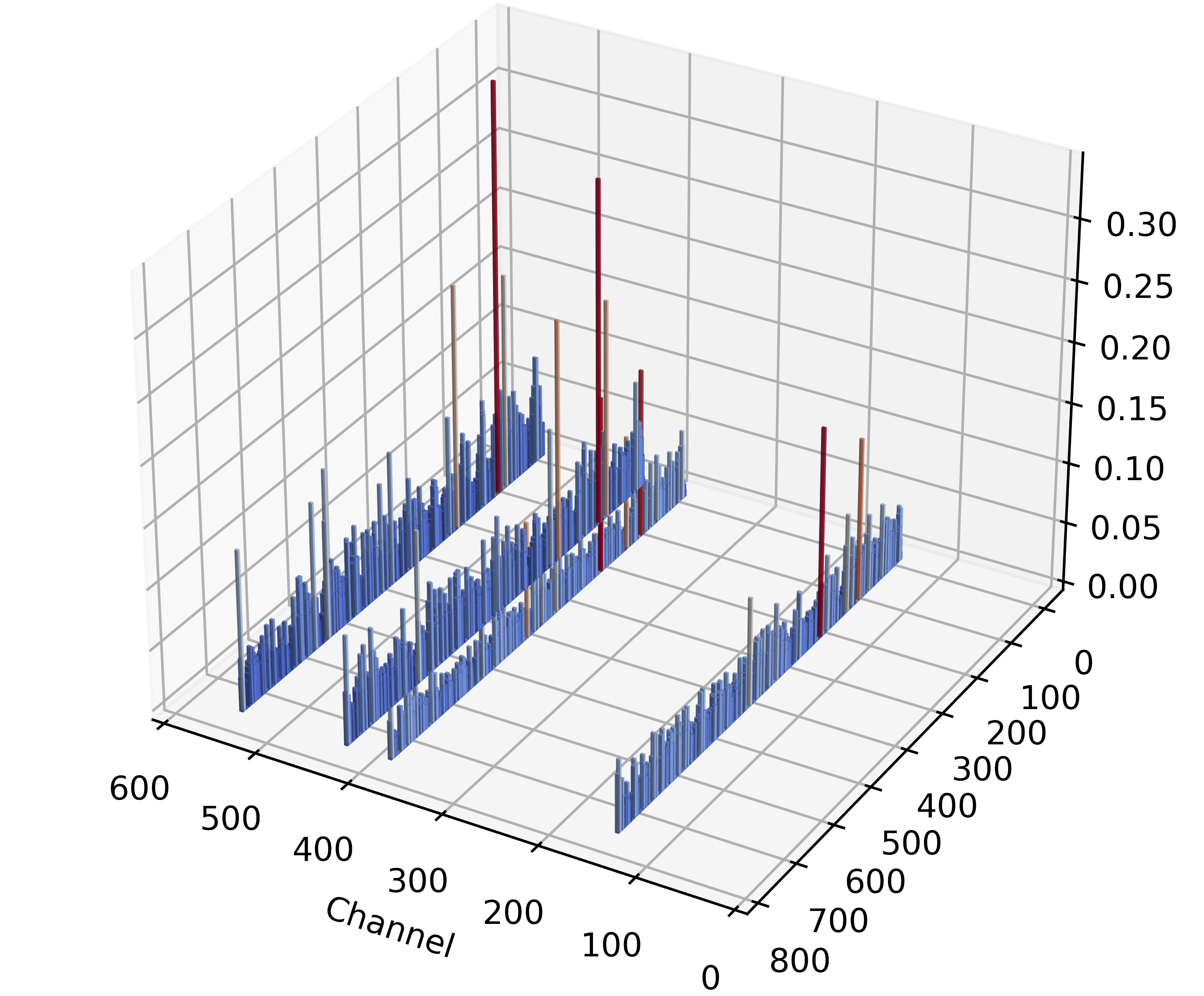}
        \label{fig:salient_weight}
    }
    \\
    \subfloat[][Channel with $\kappa=58.54$]{
        \includegraphics[width=0.5\linewidth]{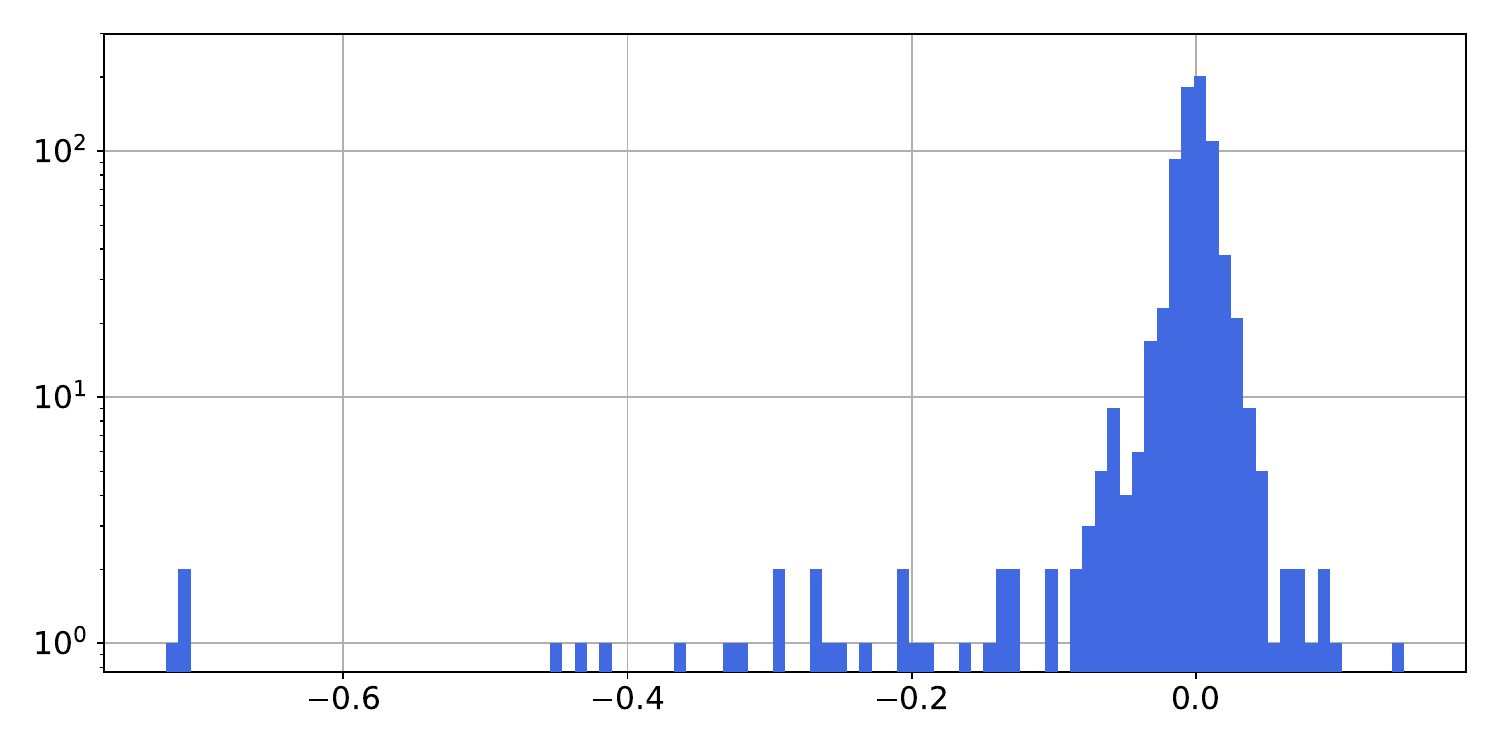}
        \label{fig:kappa_big}
    }
    \subfloat[][Channel with $\kappa=10.11$]{
        \includegraphics[width=0.5\linewidth]{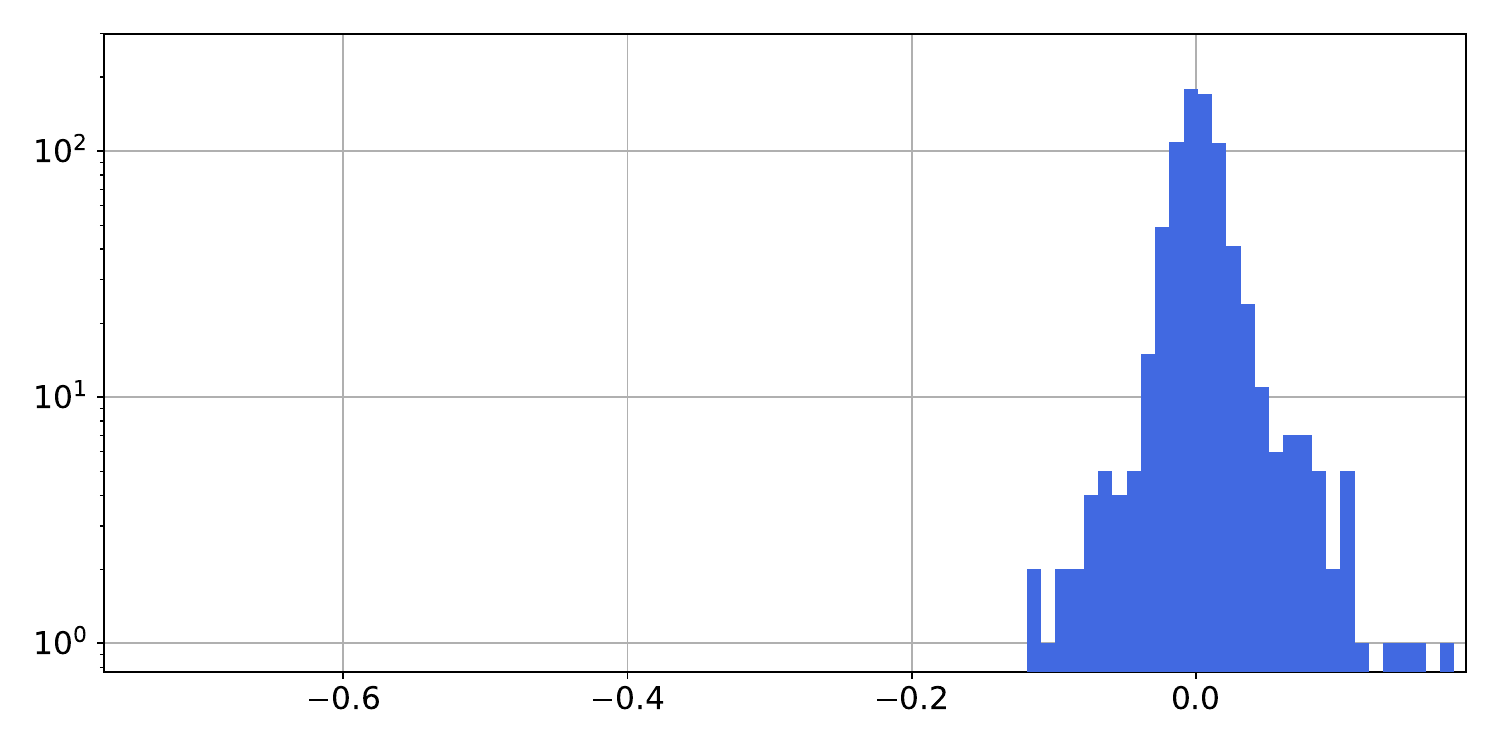}
        \label{fig:kappa_small}
    }
    \caption{Weight distribution of LDM-4 ImageNet 256x256.}
    \label{fig:weight_distribution}
\end{figure}

Studies on diffusion \cite{wu2024ptq4dit, zhao2024vidit} model find that there are some channels with significant weight values, and these channels are also crucial to the quantization error and model performance. We visualized one layer weight distribution of diffusion model in Fig. \ref{fig:layer_weight} and its corresponding outlier salient channel distribution in Fig. \ref{fig:salient_weight}. It can be found that not only the weights of different channels have a large gap in the numerical range, but some weight channels also have severe outliers. 
In Fig. \ref{fig:kappa_big} and Fig. \ref{fig:kappa_small}, we visualized two channel weight distribution. The overall weight distribution can be seen as a normal distribution plus an outlier distribution. For model quantization, direct quantization of a normal distribution is well studied \cite{liu2020reactnet, qin2023bibench}, but the presence of a small number of outliers can cause some outliers to be clamped or occupy a portion of bit width, making it difficult to quantize the main normal distribution, resulting in quantization errors. This error is greatly amplified at extremely low bit-width (2-4 bit), so we naturally hope to allocate more bit widths to these channels to ensure their performance. However, this outlier phenomenon in specific channels may occur at various layers of the model. Therefore, using traditional methods to allocate different bits to different model layers for mixed precision quantization is clearly not a good choice. But because the quantization of weights is channel-wise, we can redistribute the quantization bit width between different channels within layers to solve this problem, that is, intra-layer mixed precision quantization.

In order to determine the specific bit width allocation method, we need to quantify the significance of the outliers. $Kurtosis$ $\kappa$ can be used to quantify the "tailedness" of a real-valued variable's probability distribution \cite{decarlo1997meaningkuriosis, liu2024spinquant}. This aligns naturally with our goal, as outliers add long tails of anomalies to a normal distribution as we mentioned above. As we show in Fig. \ref{fig:kappa_big} and Fig. \ref{fig:kappa_small}, channels with more salient outlier phenomenon have higher $\kappa$ values compared to a normal distribution without outliers. Thus, we use $Kurtosis$ $\kappa$  as an indicator of the difficulty of quantizing different weight channels to find an optimal bit allocation method for each layer weight $\mathbf{W}_f$.

We hope that the model output to be as consistent as possible with the full precision model after mixed bit width assignment quantization. Since our goal is model output $\mathbf{Y}=\mathbf{X}\mathbf{W}^{\top}$, and the outliers are the maximum or minimum values in the weights, we can use the property of matrix multiplication to reduce the salient degree of outliers without loss before quantization as follows:
\begin{equation}
\begin{gathered}
    \delta_i = \sqrt{\frac{\mathrm{max}(|\mathbf{W}_i|)}{\mathrm{max}(|\mathbf{X}_i|)}}, \\
    \mathbf{Y} = (\mathbf{X} \mathrm{diag}(\delta)) \cdot (\mathrm{diag}(\delta)^{-1} \mathbf{W}^{\top}) = \hat{\mathbf{X}} \cdot \hat{\mathbf{W}}^{\top},
\end{gathered}
\end{equation}
where $\mathbf{Y}$, $\mathbf{X}$, and $\mathbf{X}$ represent the output activation, input activation, and model weights. $\delta$ is a channel-wise smooth factor that balances the quantization difficulty of weight and activation by scaling outliers closer to the normal distribution. After pre-scaled, the outlier salient channels are more smooth to be quantized. Using scaled weight $\hat{\mathbf{W}}$, we compute $\kappa$ for each weight channel and rank them accordingly. Then, we use the following optimization formula to determine the outlier salient and unsalient channels as follows:
\begin{equation}
\begin{gathered}
    {\underset{c_1,\cdots,c_g}{{\arg\min}} \ ||\mathbf{X}_f\mathbf{W}_f^{\top}, \hat{Q}(\hat{\mathbf{X}}_f)\hat{Q}(\hat{\mathbf{W}}_f|[c_1,\cdots,c_n])^{\top}||^2}, \\
     C_{N-1}=\{c_i|c_i=N-1\},\ C_{N+1}=\{c_i|c_i=N+1\}, \\
    |C_{N-1}|=|C_{N+1}|,\ |C_{N-1}|+|C_{N}|+|C_{N+1}|=n, \\
\end{gathered}
\end{equation}
where $N$ is the target average bit-width, $n$ is the channel number in weight, $\hat{Q}$ represents the quantization and de-quantization process. We apply channel-wise mixed bit-width quantization for weight as $\hat{Q}(\hat{\mathbf{W}}_f|[c_1,\cdots,c_n])$, where $c_i$ denotes the bit-width for the $i_{th}$ channel. $C$ represents the set of channel shares the same bit-width and $|C|$ stands for the number of channels in set $C$. For example, for 3-bit quantization, we assign some of the outlier salient channels to 4-bit and reassign the same number of the outlier unsalient channels to 2-bit. In this way, the average total bit-width for the layer weight is still 3-bit without adding any extra parameters.

To accelerate the optimization process, we set $k$ channels as a search group and the search region for outlier salient channel is constrained in $[0, \frac{c_{out} // k}{2}]$. We empirically set $k$ as $\frac{c_{out}}{10}$, thus the search time is $5$ times which is efficient enough for layer bit allocation.

\subsection{Time-Smoothed Relation Distillation}
For diffusion optimization, existing methods like EfficientDM~\cite{he2023efficientdm} optimize the training parameters of the quantization model by aligning the output of the full-precision (FP) model and the quantization model
\begin{equation}
    \mathcal{L}_{task} = ||\theta_f(\mathbf{x}_t, t) - \theta_q(\mathbf{x}_t, t)||^2 ,
\end{equation}
where $\theta_f$ and $\theta_q$ denotes the FP model and the quantization model respectively, $\mathbf{x}_t$ is obtained by denoising Gaussian noise $\mathbf{x}_T \sim \mathcal{N}(\mathbf{0}, \mathbf{I})$ with FP model iteratively for $T-t$ steps. 

In extremely low bit (2-4 bit) quantization, it is not sufficient to align only the final output of the model because the expressive ability of the quantization model is severely insufficient. We usually set the final projection layer of the quantization model to 8-bit and the layers before the project layer to target low bit~\cite{wang2024quest, he2023efficientdm, huang2024tfmq}. In this way, we do not directly perceive the part where the quantization information is severely missing. Therefore, we can give more fine-grained guidance to the extremely low bit quantization model by distilling the model feature layer before the last project layer
\begin{equation}
    \mathcal{L}_{dis} = \mathcal{D}(\mathbf{F}_f, \mathbf{F}_q),
\end{equation}
where $\mathbf{F}_f$ and $\mathbf{F}_q$ denote the feature map before the last project layer of FP model and quantization model respectively. $\mathcal{D}$ is a metric to measure the distance of the two feature maps.

\begin{figure}[t]
    \centering
    \subfloat[][$T=20$]{
        \includegraphics[width=1.0\linewidth]{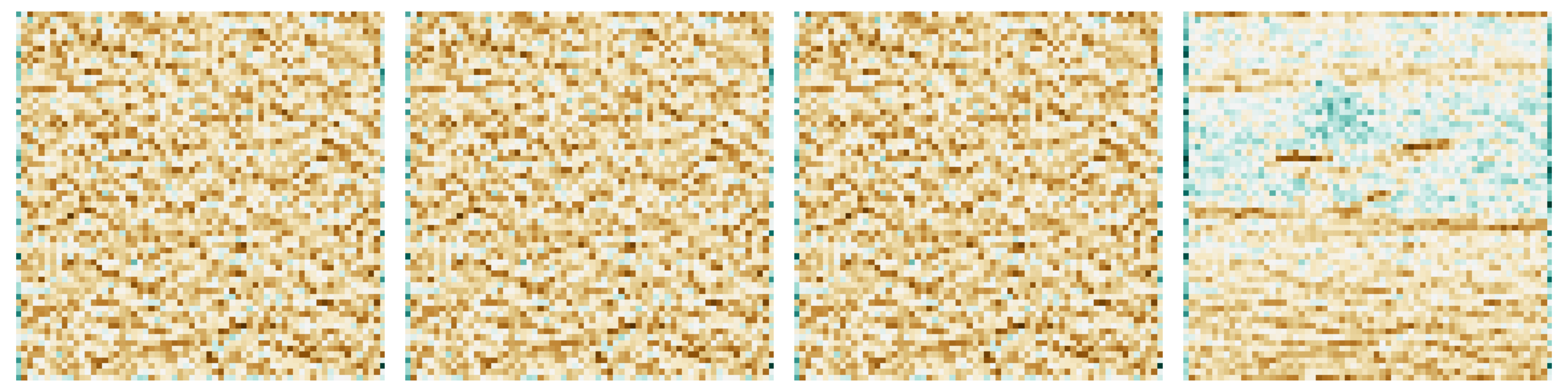}
        \label{}
    }
    \\
    \subfloat[][$T=15$]{
        \includegraphics[width=1.0\linewidth]{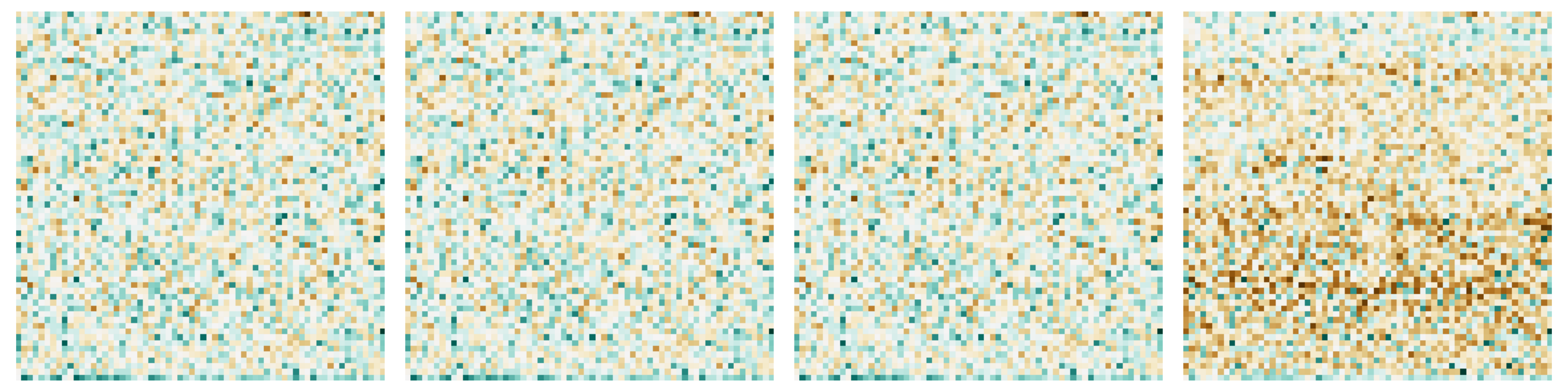}
        \label{}
    }
    \caption{Visualization of feature map from timesetp (Left) $T$, (Left-Mid) $T-1$, (Right-Mid) $T-2$, (Right) $T-10$.}
    \label{fig:timestep}
\end{figure}

However, features at extremely low bit quantization present high discretization (e.g., only 16 values for 4-bit activation quantization) \cite{martinez2020trainingbinary}. This discretized feature shows a high degree of unrobustness in numerical value, and directly distilling it has poor effect or even impairs the normal training of the quantization model~\cite{zheng2024binarydm, qin2022bibert}. Moreover, due to the unique iterative denoising process of the diffusion model~\cite{ho2020ddpm, song2020ddim}, the outliers of the features will continue to add up with denoising, further amplifying this unrobustness. Typically we can use additional projection heads to map them to a uniform space or to regularize it numerically \cite{yang2022cirkd, yang2023online, feng2024rdd, yang2024clipkd}. However, projection heads will bring additional training parameters, and performing regularization cannot solve the problem of error accumulation in iterative calculation. We find that the features of the diffusion model are highly correlated with time steps and show similarity in some time steps. In Fig. \ref{fig:timestep}, we find that the features on consecutive time steps are highly similar, while the time steps far apart are quite different. The similarity of features indicates that the difference of consecutive time steps in denoising trajectory is quite small, so we can alleviate the feature unrobustness at different time steps by fusing the intermediate features of multiple consecutive steps. Instead of forcing the highly discretized quantization model to strictly learn the denoising trajectory of each step of FP model, but to learn the denoising trajectory of successive multiple steps.

Therefore, we use the intermediate features of $N$ consecutive steps as the smoothed feature representation for distillation. For $T$ timestep optimization, we rewrite the distillation formula as
\begin{equation}
\begin{gathered}
    \hat{\mathbf{F}}_f = \sum_{t=0}^{N} \mathbf{F}_{T-t, f}, \
    \hat{\mathbf{F}}_q = \sum_{t=0}^{N} \mathbf{F}_{T-t, q}, \\
    \mathcal{L}_{dis} = \mathcal{D}(\hat{\mathbf{F}}_f || \hat{\mathbf{F}}_q),
\end{gathered}
\label{eq:time_smooth}
\end{equation}
where $\mathbf{F}_{T-t, f}$ and $\mathbf{F}_{T-t, q}$ denotes the last feature map in time step $T-t$ from FP model and quantization model respectively.
%

\begin{figure}
    \centering
    \subfloat[][Original feature map]{
        \includegraphics[width=1.0\linewidth]{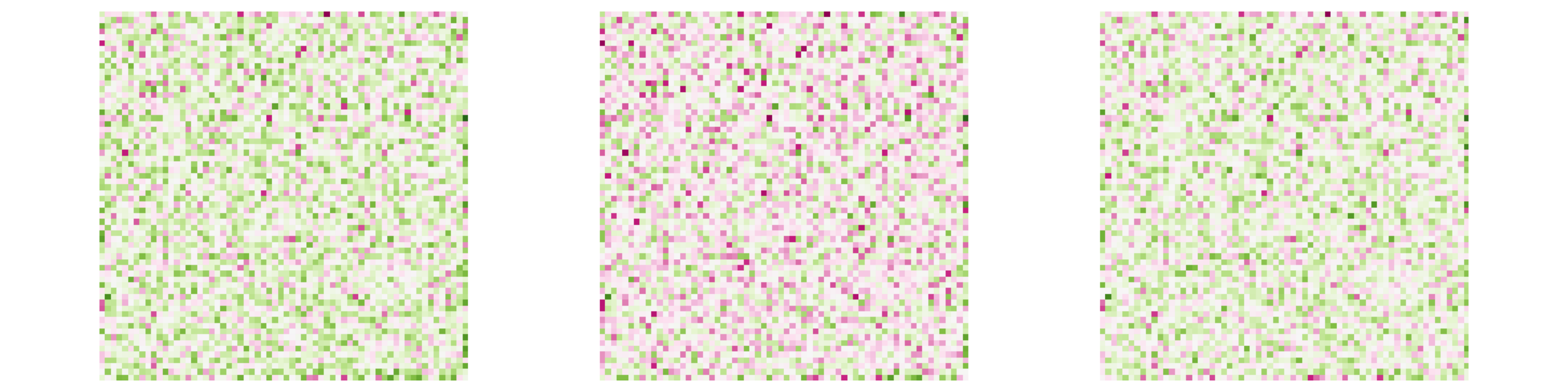}
        \label{fig:mse_mismatch}
    }
    \\
    \subfloat[][Cosine similarity map]{
        \includegraphics[width=1.0\linewidth]{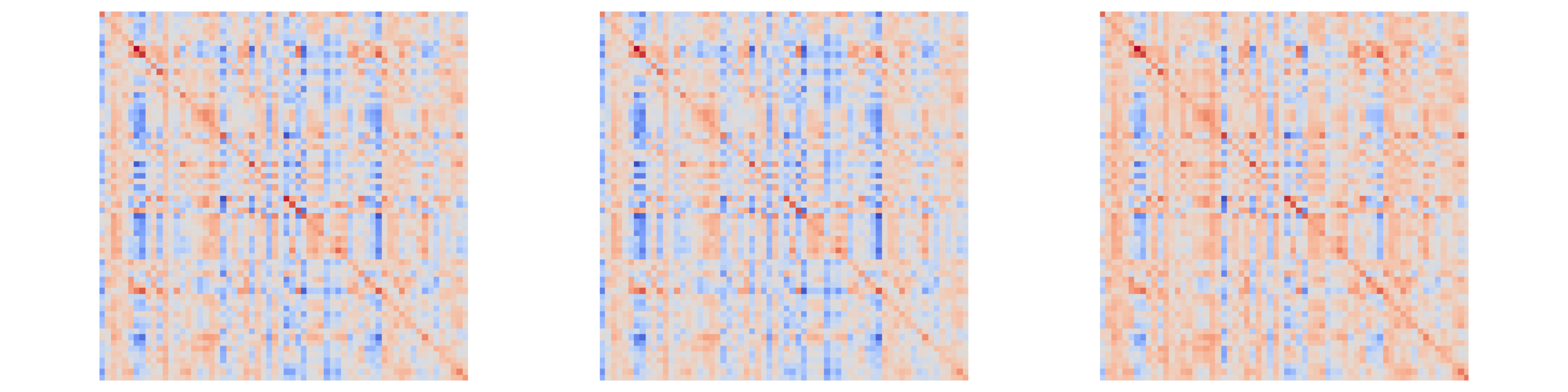}
        \label{fig:kl_mismatch}
    }
    \caption{Visualization of different activation maps of (Left) FP model, (Mid) well-trained quantization model, (Right) un-trained quantization model.}
    \label{fig:mse_with_kl}
\end{figure}

In Fig. \ref{fig:mse_mismatch}, we visualized feature maps between FP model, well-trained quantization model, and un-trained quantization model. Although time-smoothed feature improves the robustness of quantized feature, there is still a mismatch in numerical expression between well-trained model and FP model. This is blamed on the difference between the discrete latent space of quantized features and continuous latent space of FP features. Thus, any metrics such as L2 loss that numerically align $\hat{\mathbf{F}}_f$ and $\hat{\mathbf{F}}_q$ cannot avoid this difference between spaces. Therefore, we propose to use relation distillation to replace the strict numerical alignment by learning the feature similarity relation between $\hat{\mathbf{F}}_f$ and $\hat{\mathbf{F}}_q$. In Fig \ref{fig:kl_mismatch}, we transfer the numerical relationship between discrete latent space and continuous latent space to the feature similarity relationship inside each space which unifies the distillation goal into the feature similarity space. This successfully solved the numerical mismatch. Formally speaking, for $\hat{\mathbf{F}} \in \mathbb{R}^{h\times w\times c}$ and to simplify writing, we reshape it as $\hat{\mathbf{F}} \in \mathbb{R}^{s\times c}$, where $s=h\times w$. For $i_{th}$ feature representation $\hat{\mathbf{F}}^i$, we can calculate its cosine similarity distribution with each position feature representation $\hat{\mathbf{S}}^i = \hat{\mathbf{F}}^i\hat{\mathbf{F}}^{\top} \in \mathbb{R}^{s}$. Thus, the relation distillation metric is
\begin{equation}
    \mathcal{L}_{dis} = \sum^{s}_{i=1} \mathcal{D}_{kl}(\hat{\mathbf{S}}^i_f || \hat{\mathbf{S}}^i_q),
\end{equation}
where $\hat{\mathbf{S}}^i_f$ and $\hat{\mathbf{S}}^i_q$ denotes time-smoothed feature similarity distribution from FP model and quantization model respectively. $\mathcal{D}_{kl}(\cdot || \cdot)$ stands for the Kullback-Leibler (KL) divergence between two distributions. We utilize KL divergence instead of L2 Loss here. Because L2 Loss can only perceive a single representation but KL divergence can perceive the information of the whole feature map. The overall optimization target is
\begin{equation}
    \mathcal{L}_{total} = \mathcal{L}_{target} + \lambda \mathcal{L}_{dis}.
\end{equation}

%% file: sec/5_experiment.tex
\section{Experiment}

\begin{table}[!htb]
    \centering
    \fontsize{9}{13}\selectfont
    \begin{tabular}{lcccccc}
    \hline
        Method & \makecell{Bit \\ (W/A)} & \makecell{Size \\ (MB)} & IS $\uparrow$ & FID $\downarrow$ 
        & sFID $\downarrow$ & \makecell{Precision $\uparrow$ \\ (\%)} \\ \hline
        FP & 32/32 & 1529.7 & 364.73 & 11.28 & 7.70 & 93.66 \\ \hline
        PTQ-D & 3/6 & 144.5 & 162.90 & 17.98 & 57.31 & 63.13 \\
        TFMQ & 3/6 & 144.5 & 174.31 & 15.90 & 40.63 & 67.42 \\
        QuEST & 3/6 & 144.6 & 194.32 & 14.32 & 31.87 & 72.80 \\
        EfficientDM & 3/6 & 144.6 & 299.63 & 7.23 & 8.18 & 86.27 \\
        HAWQ-V3 & 3/6 & 144.6 & \underline{303.79} & \underline{6.94} & \underline{8.01} & \underline{87.76} \\
        \textbf{MPQ-DM} & 3/6 & 144.6 & \textbf{306.33} & \textbf{6.67} & \textbf{7.93} & \textbf{88.65} \\ \hline
        PTQ-D & 3/4 & 144.5 & 10.86 & 286.57 & 273.16 & 0.02 \\
        TFMQ & 3/4 & 144.5 & 13.08 & 223.51 & 256.32 & 0.04 \\
        QuEST & 3/4 & 175.5 & 15.22  & 202.44 & 253.64 & 0.04 \\
        EfficientDM & 3/4 & 144.6 & 134.30  & 11.02 & 9.52 & 70.52 \\
        HAWQ-V3 & 3/4 & 144.6 & \underline{152.61} & \underline{8.49} & \underline{9.26} & \underline{75.02} \\
        \textbf{MPQ-DM} & 3/4 & 144.6 & \textbf{197.43} & \textbf{6.72} & \textbf{9.02} & \textbf{81.26} \\ \hline
        PTQ-D & 2/6 & 96.7 & 70.43 & 40.29 & 35.70 & 43.79 \\
        TFMQ & 2/6 & 96.7 & 77.26 & 36.22 & 33.05 & 45.88 \\
        QuEST & 2/6 & 96.8 & 86.83 & 32.37 & 31.58 & 47.74 \\
        EfficientDM & 2/6 & 96.8 & 69.64 & 29.15 & 12.94 & 54.70 \\
        HAWQ-V3 & 2/6 & 96.8 & 88.25 & 22.73 & 11.68 & 57.04 \\
        MPQ-DM & 2/6 & 96.8 & \underline{102.51} & \underline{15.89} & \underline{10.54} & \underline{67.74} \\
        \textbf{MPQ-DM}$^+$ & 2/6 & 101.6 & \textbf{136.35} & \textbf{11.00} & \textbf{9.41} & \textbf{72.84} \\ \hline
        PTQ-D & 2/4 & 96.7 & 9.25 & 336.57 & 288.42 & 0.01 \\
        TFMQ & 2/4 & 96.7 & 12.76 & 300.03 & 272.64 & 0.03 \\
        QuEST & 2/4 & 127.7 & 14.09 & 285.42 & 270.12 & 0.03 \\
        EfficientDM & 2/4 & 96.8 & 25.20 & 64.45 & 14.99 & 36.63 \\
        HAWQ-V3 & 2/4 & 96.8 & 33.21 & 52.63 & 14.00 & 42.95 \\
        MPQ-DM & 2/4 & 96.8 & \underline{43.95} & \underline{36.59} & \underline{12.20} & \underline{52.14} \\
        \textbf{MPQ-DM}$^{+}$ & 2/4 & 101.6 & \textbf{60.55} & \textbf{27.11} & \textbf{11.47} & \textbf{57.84} \\ \hline
    \end{tabular}
    \caption{Performance comparisons of fully-quantized LDM-4 models on ImageNet 256$\times$256. Best results are in bold and second best are in underlined.}
    \label{tab:imagenet}
\end{table}

\begin{table}[t]
    \centering
    \setlength{\tabcolsep}{0.8mm}
    \begin{tabular}{lcccccc}
    \hline
        Method & \makecell{Bit \\ (W/A)} & Size (MB) & FID $\downarrow$ & sFID $\downarrow$ & \makecell{Precision $\uparrow$ \\ (\%)} \\ \hline
        FP & 32/32 & 1045.4 & 7.39 & 12.18 & 52.04 \\ \hline
        PTQ-D & 3/6 & 98.3 & 113.42 & 43.85 & 10.06\\
        TFMQ & 3/6 & 98.3 & 26.42 & 30.87 & 38.29 \\
        QuEST & 3/6 & 98.4 & 21.03 & 28.75 & 40.32 \\
        EfficientDM & 3/6 & 98.4 & \underline{13.37} & \underline{16.14} & \underline{44.55} \\
        \textbf{MPQ-DM} & 3/6 & 98.4 & \textbf{11.58} & \textbf{15.44} & \textbf{47.13} \\ \hline
        PTQ-D & 3/4 & 98.3 & 100.07 & 50.29 & 11.64 \\
        TFMQ & 3/4 & 98.3 & 25.74 & 35.18 & 32.20 \\
        QuEST & 3/4 & 110.4 & \underline{19.08} & 32.75 & \underline{40.64} \\
        EfficientDM & 3/4 & 98.4 & 20.39 & \underline{20.65} & 38.70 \\
        \textbf{MPQ-DM} & 3/4 & 98.4 & \textbf{14.80} & \textbf{16.72} & \textbf{43.61} \\ \hline
        PTQ-D & 2/6 & 65.7 & 86.65 & 53.52 & 10.27 \\
        TFMQ & 2/6 & 65.7 & 28.72 & 29.02 & 34.57 \\
        QuEST & 2/6 & 65.7 & 29.64 & 29.73 & 34.55 \\
        EfficientDM & 2/6 & 65.7 & 25.07 & 22.17 & 34.59 \\
        MPQ-DM & 2/6 & 65.7 & \underline{17.12} & \underline{19.06} & \underline{40.90} \\
        \textbf{MPQ-DM}$^+$ & 2/6 & 68.9 & \textbf{16.54} & \textbf{18.36} & \textbf{41.80} \\ \hline
        PTQ-D & 2/4 & 65.7 & 147.25 & 49.97 & 9.26 \\
        TFMQ & 2/4 & 65.7 & 25.77 & 36.74 & 32.86 \\
        QuEST & 2/4 & 77.7 & 24.92 & 36.33 & 32.82 \\
        EfficientDM & 2/4 & 65.7 & 33.09 & 25.54 & 28.42 \\
        MPQ-DM & 2/4 & 65.7 & \underline{21.69} & \underline{21.58} & \underline{38.69} \\
        \textbf{MPQ-DM}$^+$ & 2/4 & 68.9 & \textbf{20.28} & \textbf{19.42} & \textbf{38.92} \\ \hline
    \end{tabular}
    \caption{Unconditional image generation results of LDM-4 models
 on LSUN-Bedrooms 256$\times$256. 
    }
    \label{tab:lsun_bedroom}
\end{table}

\subsection{Experiment Settings}
We conduct experiments on commonly used datasets LSUN-Bedrooms 256×256, LSUN-Churches 256×256 \cite{yu2015lsun}, and ImageNet 256×256 \cite{deng2009imagenet} for both unconditional and conditional image generation tasks on LDM models. We also conduct text-to-image generation task on Stable Diffusion \cite{rombach2022ldm}. We use IS~\cite{salimans2016is_metric}, FID \cite{heusel2017fid}, sFID \cite{nash2021sfid} and Precision to evaluate LDM performance. For Stable Diffusion, we use CLIP Score \cite{hessel2021clipscore} for evaluation. To cope with the extreme expressivity degradation under 2bit quantization, we allocate an additional 10\% number of channels for 2bit during the search process of OMQ, named MPQ-DM$^+$. This results in only a 0.6\% increase in model size compared with FP model. We compare our MPQ-DM with baseline EfficientDM \cite{he2023efficientdm} and layer-wise mixed precision HAWQ-v3 \cite{yao2021hawqv3} and other PTQ-based methods \cite{he2024ptqd, huang2024tfmq, wang2024quest} which possess similar time consumption. Details can be found in Appendix.

\subsection{Experiment Results}

\textbf{Class-conditional Generation.}
We conduct conditional generation experiment on ImageNet 256$\times$256 dataset, focusing on LDM-4. Results in Table \ref{tab:imagenet} show that MPQ-DM greatly outperforms existing methods on all bit settings. MPQ-DM generally performs better than the layer-wise approach HAWQ-v3, demonstrating the necessity of mixed precision quantization within layers. MPQ-DM W3A4 model even surpasses FP model on FID. In W2A4 setting, PTQ-based methods fail to generate images, while EfficientDM performs poorly. MPQ-DM greatly improves baseline with a notable 27.86 decrease in FID. MPQ-DM$^+$ even further leads to 9.48 decrease in FID using only 4.8 MB additional model size.

\textbf{Unconditional Generation.}
We conduct unconditional generation experiment on LSUN-Bedrooms dataset over LDM-4 and LSUN-Churches dataset over LDM-8 with 256$\times$256 resolution. In Table \ref{tab:lsun_bedroom} and Table \ref{tab:lsun_church}, MPQ-DM still outperforms all other existing methods under all bit settings. For LSUN-Bedrooms dataset, we achieved FID decrease of 5.59 on W3A4, 8.53 on W2A6, and even 12.81 on W2A4 setting compared with baseline. Under W2A4 setting, we are the first method pushing sFID under 20 which leads to 6.12 decrease compared with baseline.

\textbf{Text-to-image Generation.}
We conduct text-to-image generation experiment on randomly selected 10k COCO2014 validation set prompts over Stable Diffusion v1.4 model with 512$\times$512 resolution. In Table \ref{tab:sd}, our method achieves better performance over baseline and SOTA PTQ methods. In W3A4 and W2A6 settings, we achieve over 0.3 CLIP Score improvement. MPQ-DM$^+$ even further achieves 1.79 improvement in CLIP Score with only 10.2 MB additional model size. 

\begin{table}[t]
    \centering
    \setlength{\tabcolsep}{0.8mm}
    \begin{tabular}{lcccccc}
    \hline
        Method & \makecell{Bit \\ (W/A)} & Size (MB) & FID $\downarrow$ & sFID $\downarrow$ & Precision $\uparrow$ \\ \hline
        FP & 32/32 & 1125.2 & 5.55 & 10.75 & 67.43 \\ \hline
        PTQ-D & 3/6 & 106.0 & 59.43 & 40.26 & 13.37 \\
        TFMQ & 3/6 & 106.0 & 13.53 & 22.10 & 62.74 \\
        QuEST & 3/6 & 106.1 & 22.19 & 32.79 & 60.73 \\
        EfficientDM & 3/6 & 106.1 & \underline{9.53} & \underline{13.70} & \underline{62.92} \\
        \textbf{MPQ-DM} & 3/6 & 106.1 & \textbf{9.28} & \textbf{13.37} & \textbf{63.73} \\ \hline
        PTQ-D & 3/4 & 106.0 & 77.08 & 49.63 & 10.25 \\
        TFMQ & 3/4 & 106.0 & 35.51 & 48.59 & 55.32 \\
        QuEST & 3/4 & 122.4 & 40.74 & 53.63 & 52.78 \\
        EfficientDM & 3/4 & 106.1 & \underline{15.59} & \underline{18.16} & \underline{57.92} \\
        \textbf{MPQ-DM} & 3/4 & 106.1 & \textbf{14.08} & \textbf{16.91} & \textbf{59.68} \\ \hline
        PTQ-D & 2/6 & 70.9 & 63.38 & 46.63 & 12.14 \\
        TFMQ & 2/6 & 70.9 & 25.51 & 35.83 & 54.75 \\
        QuEST & 2/6 & 70.9 & 23.03 & 35.13 & 56.90 \\
        EfficientDM & 2/6 & 70.9 & 16.98 & 18.18 & 57.39 \\
        MPQ-DM & 2/6 & 70.9 & \underline{15.61} & \underline{17.44} & \underline{59.03} \\
        \textbf{MPQ-DM}$^+$ & 2/6 & 74.4 & \textbf{13.38} & \textbf{15.59} & \textbf{61.00} \\ \hline
        PTQ-D & 2/4 & 70.9 & 81.95 & 50.66 & 9.47 \\
        TFMQ & 2/4 & 70.9 & 51.44 & 64.07 & 42.25 \\
        QuEST & 2/4 & 86.9 & 50.53 & 63.33 & 45.86 \\
        EfficientDM & 2/4 & 70.9 & 22.74 & 22.55 & 53.00 \\
        MPQ-DM & 2/4 & 70.9 & \underline{21.83} & \underline{21.38} & \underline{53.99} \\
        \textbf{MPQ-DM}$^+$ & 2/4 & 74.4 & \textbf{16.91} & \textbf{18.57} & \textbf{58.04} \\ \hline
    \end{tabular}
    \caption{Unconditional image generation results of LDM-8 models
 on LSUN-Churches 256$\times$256. 
    }
    \label{tab:lsun_church}
\end{table}

\begin{table}[!htp]
    \centering
    \setlength{\tabcolsep}{2.2mm}
    \begin{tabular}{lccc}
    \hline
        Method & Bit (W/A) & Size (MB) & CLIP Score $\uparrow$ \\ \hline
        FP & 32/32 & 3279.1 & 31.25 \\ \hline
        QuEST & 3/4 & 332.9 & 26.55 \\
        EfficientDM & 3/4 & 309.8 & \underline{26.63} \\
        \textbf{MPQ-DM} & 3/4 & 309.8 & \textbf{26.96} \\ \hline
        QuEST & 2/6 & 207.4 & 22.88 \\
        EfficientDM & 2/6 & 207.4 & 22.94 \\
        MPQ-DM & 2/6 & 207.4 & \underline{23.23} \\
        \textbf{MPQ-DM}$^+$ & 2/6 & 217.6 & \textbf{25.02} \\ \hline
    \end{tabular}
    \caption{Text-to-image generation results (512$\times$512) of Stable Diffusion v1.4 using 10k COCO2014 validation set prompts. 
    } 
    \label{tab:sd}
\end{table}

\subsection{Ablation Study}
\textbf{Component Study.} In Table \ref{tab:ablation_method}, we perform comprehensive ablation studies on LDM-4 ImageNet 256$\times$256 model to evaluate the effectiveness of each proposed component. Our proposed OMQ solves the existing layer-wise bit allocation methods from the perspective of quantization, allocating more bit width to the channels with salient outlier phenomenon within layer while the total average bit width is unchanged. This intra-layer mixed-precision quantization method greatly improves the performance of baseline, gaining IS increases of 58.88. In addition, TSD improves the robustness in the distillation process from the perspective of model optimization, and also achieves a certain improvement. Through the parallel improvement of the two perspectives of quantization and optimization, MPQ-DM achieves state-of-the-art quantization performance.
\begin{table}[h]
    \centering
    \fontsize{9}{13}\selectfont
    \setlength{\tabcolsep}{1.8mm}
    \begin{tabular}{lccccc}
    \hline
         Method & \makecell{Bit \\ (W/A)} & IS $\uparrow$ & FID $\downarrow$ & sFID $\downarrow$ & Precision $\uparrow$ \\ \hline
         PTQD & 3.38 & 10.86 & 286.57 & 237.16 & 0.05 \\ \hline
         Baseline & 3.05 & 134.30 & 11.02 & 9.52 & 70.52 \\ \hline 
         +OMQ & 4.01 & 193.18 & 6.91 & 9.12 & 80.77 \\
         +TSD & 3.10 & 135.91 & 10.38 & 9.38 & 72.21 \\
         MPQ-DM & 4.05 & \textbf{197.43} & \textbf{6.72} & \textbf{9.02} & \textbf{81.26}\\ \hline
    \end{tabular}
    \caption{Ablation study on proposed methods. 
    }
    \label{tab:ablation_method}
\end{table}

\textbf{Outlier Selection Method Study.} In Table \ref{tab:ablation_channel}, we study different outlier salient channel selection methods in mixed precision quantization on LDM-4 ImageNet 256$\times$256 model. We find that even randomly selecting some channels to higher or lower bits leads to a certain performance gain. This indicates that in extremely low bit quantization, the gain brought by increasing the bit of some channels is far greater than the impact brought by decreasing some channels, which proves the necessity of mixed quantization. While there are some gains in selecting channels randomly or from the head and tail of weights, our outlier selection method based on $Kurtosis$ achieves the most significant performance improvement. This shows that $Kurtosis$ selects the outlier salient channels with the most significant performance improvement, which proves the effectiveness of our $ Kurtosis$-based channel selection method.
\begin{table}[!tb]
    \centering
    \setlength{\tabcolsep}{2.2mm}
    \begin{tabular}{lccccc}
    \hline
         Method & \makecell{Bit \\ (W/A)} & FID $\downarrow$ & sFID $\downarrow$ & Precision $\uparrow$ \\ \hline
         Baseline & 3/4 & 11.08 & 22.02 & 75.86 \\ \hline 
         Random & 3/4 & 9.69 & 22.23 & 79.63 \\
         Head-tail & 3/4 & 9.44 & 21.71 & 79.95 \\
         $Kurtosis \ \kappa$ & 3/4 & \textbf{9.05} & \textbf{21.51} & \textbf{80.60} \\ \hline
    \end{tabular}
    \caption{Ablation study on outlier selection function. 
    We sample 10k samples for evaluation.}
    \label{tab:ablation_channel}
\end{table}

\textbf{Distillation Metrics Study.} In Table \ref{tab:ablation_distill}, we study different distillation metrics used in Eq. \ref{eq:time_smooth} on LDM-4 ImageNet  model. We compare with without distillation to valid different metrics. Using L2 Loss to align $\hat{\mathbf{F}}_f$ and $\hat{\mathbf{F}}_q$ only shows little improvement on sFID, but decreases FID and Precision. This indicates that the discrete features of and continuous features cannot be well aligned by numerical values directly, which leads to negative optimization. However, our proposed relation distillation can transfer all features into a unified similarity space. This breaks the difference between the two latent spaces and improves model performance.
\begin{table}[!htb]
    \centering
    \fontsize{9}{12}\selectfont
    \setlength{\tabcolsep}{2.0mm}
    \begin{tabular}{lccccc}
    \hline
         Method & \makecell{Bit \\ (W/A)} & FID $\downarrow$ & sFID $\downarrow$ & Precision $\uparrow$ \\ \hline
         w/o Distillation & 3/4 & 9.12 & 21.59 & 81.15 \\ \hline 
         L2 Loss & 3/4 & 9.13 & 21.47 & 80.76 \\
         Relation Distillation & 3/4 & \textbf{9.10} & \textbf{21.40} & \textbf{81.25} \\ \hline
    \end{tabular}
    \caption{Ablation study on distillation metrics. 
    We sample 10k samples for evaluation.}
    \label{tab:ablation_distill}
\end{table}

%% file: sec/6_conclusion.tex
\section{Conclusion}
In this paper, we have proposed MPQ-DM, a Mixed-Precision Quantization method for extremely low bit diffusion quantization. To address severe model quantization error caused by outlier salient weight channel, we have proposed Outlier-Driven Mixed Quantization to apply optimized intra-layer mixed-precision bit-width allocation that automatically resolves performance degradation introduced by the outlier. To robustly learn representations across time steps, we have constructed a Time-Smoothed Relation Distillation scheme to obtain feature representations that are more suitable for quantitative model learning. Our extensive experiments have demonstrated the superiority of MPQ-DM over baseline and other previous PTQ-based methods.

%% file: sec/supp.tex
\section{Experiment Settings}

\subsection{Experimental Hardware}
All our experiments were conducted on a server with Intel(R) Xeon(R) Gold 5318Y 2.10@GHz CPU and NVIDIA A800 80GB GPU.

\subsection{Models and Dataset}
We perform comprehensive experiments encompassing unconditional image generation, class-conditional image generation, and text-conditional image generation tasks on two diffusion models: latent-space diffusion model LDM and Stable Diffusion v1.4. For LDM, our investigations spanned multiple datasets, including LSUN-Bedrooms, LSUN-Churches, and ImageNet, all with a resolution of 256$\times$256. Furthermore, we employ Stable Diffusion for text-conditional image generation on randomly sampled 10k COCO2014 validation set prompts with a resolution of 512$\times$512. This diverse set of experiments, conducted on different models, datasets, and tasks, allows us to validate the effectiveness of our MPQ-DM comprehensively.

\subsection{Pipeline and Hyperparameters}
We follow EfficientDM to perform quantization-aware low-rank fine-tuning to quantization diffusion models. We perform Outlier-Driven Mixed Quantization at the beginning of quantization process to decide bit-width for different channels. We apply Time-Smoothed Relation Distillation during the optimization process to improve feature robustness. For LDM models training process, we fine-tune LoRA weights and quantization parameters for 16K iterations with a batchsize of 4 same as EfficientDM. And we use the same learning rate and optimizer as in EfficientDM. We set hyperparameter $\alpha=100$ in Time-Smoothed Relation Distillation. For Stable Diffusion training process, we fine-tune parameters for 30K iterations with a batchsize of 2 and set $\alpha=1$ while all other settings remain the same with LDM models.

\subsection{Evaluation}
To assess the generation quality of the LDM models, we utilize several evaluation metrics, including Inception Score (IS), Fréchet Inception Distance (FID), Sliding Fréchet Inception Distance (sFID), and Precision. In each evaluation, we randomly generate 50,000 samples from the model and compute the metrics using reference batches. The reference batches used to evaluate FID and sFID contain all the corresponding datasets, while only 10,000 images were extracted when Precision is calculated. We recorded FID, sFID, and Precision for all tasks and additional IS for ImageNet. Because the Inception Score is not a reasonable metric for datasets that have significantly different domains and categories from ImageNet. These metrics are all evaluated using ADM’s TensorFlow evaluation suite. For Stable Diffusion, we use CLIP Score for text-image alignment as it includes additional text information.

\section{More Abaltion Study}
\subsection{Smooth Step Study}
In Table \ref{tab:ablation_timestep}, we evaluate different $N$ values used in Eq. \ref{eq:time_smooth} for Time-Smoothed Relation Distillation. $N=0$ denotes direct distillation without time smooth. It can be found that all different $N$ consecutive steps for time smooth can improve model performance. This shows our time-smoothed feature can improve the robustness of intermediate representations, obtaining better optimization results. In our experiment, we use $N=1$ since it achieves good performance.
\begin{table}[!htb]
    \centering
    \begin{tabular}{c|ccccc}
    \hline
         $N$ & 0 & \underline{1} & 2 & 3 & 4 \\ \hline
         sFID $\downarrow$ & 23.37 & \textbf{22.42} & 23.19 & 22.76 & 22.61 \\ \hline 
    \end{tabular}
    \caption{Ablation study on different $N$ used in Time-Smoothed Relation Distillation. Experiment conducted on LDM-4 ImageNet 256$\times$256 model under W3A4 quantization. We sample 10k samples for evaluation.}
    \label{tab:ablation_timestep}
\end{table}

\section{More Visualization}
\subsection{Visualization of Target Channel Weight}
\begin{figure*}
    \centering
    \includegraphics[width=1.0\textwidth]{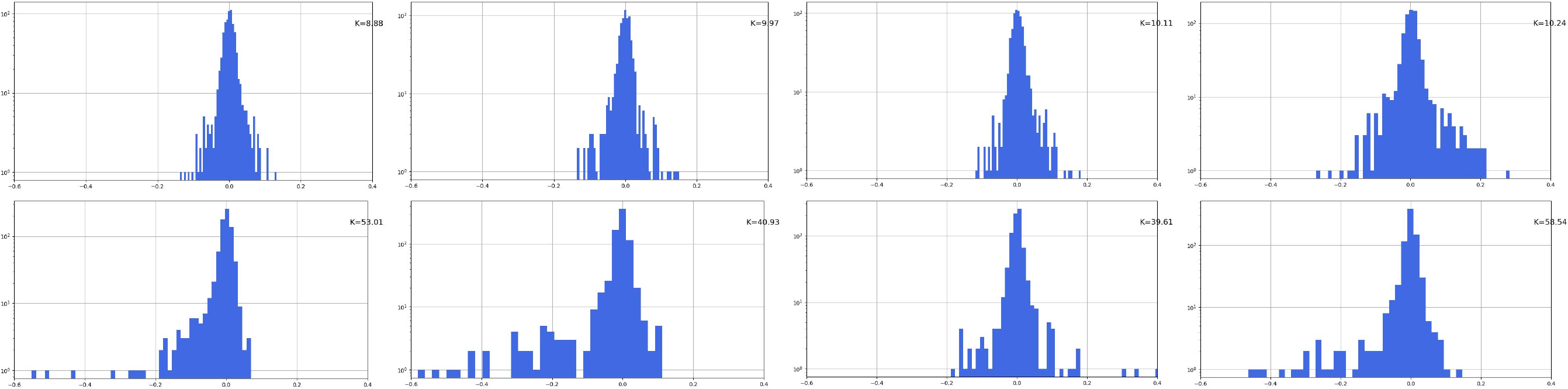}
    \label{}
    \caption{More visualization of different channel weight distribution. Upper row with lower $Kurtosis$, lower row with higher $Kurtosis$.}
    \label{fig:channel_weight}
\end{figure*}

In Fig. \ref{fig:channel_weight}, we present here different channel weight distribution with different $Kurtosis$ magnitude. Upper row channels have lower $Kurtosis$, and their distribution presents a more concentrated normal distribution with almost no outliers. While lower row channels have higher $Kurtosis$. Their distribution not only follows a concentrated normal distribution but also has salient outliers. This indicates that $kurtosis$ is a good measure of the significance of outliers in weight distribution.

\subsection{Visualization of Different Layer Weight}
\begin{figure*}
    \centering
    \includegraphics[width=1.0\textwidth]{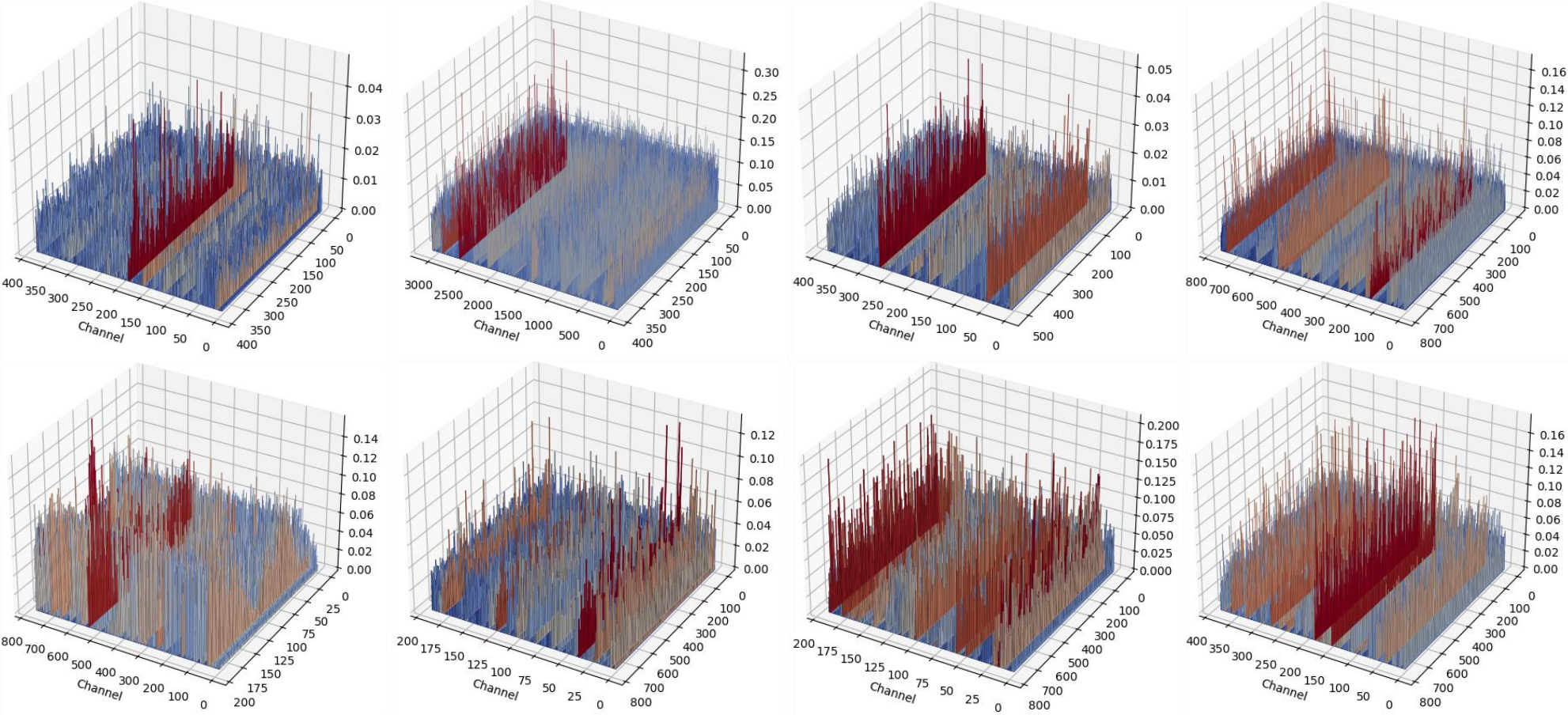}
    \label{}
    \caption{More visualization of different layer weight distribution.}
    \label{fig:weight_3d}
\end{figure*}

In Fig. \ref{fig:weight_3d}, we present here different layers overall weight distribution. It can be observed that the numerical distribution between different channels in weights is significantly different, and there are always some outliers in the weights of each layer. This proves the necessity of using intra-layer mixed precision quantization to address outliers in different layers

\subsection{Additional Random Samples}
We further showcase more random generation results on various datasets, with
unconditional generation on LSUN-Bedrooms, LSUN-Churches, class-conditional generation on ImageNet, and text-conditional generation on Stable Diffusion. Overall, MPQ-DM exhibits the best generation performance across datasets. In contrast, the Baseline tends to exhibit noticeable exposure errors and lacks detailed textures.

\begin{figure*}
\begin{minipage}{\textwidth}
    \centering
    \subfloat[][Baseline]{
        \includegraphics[width=0.5\textwidth]{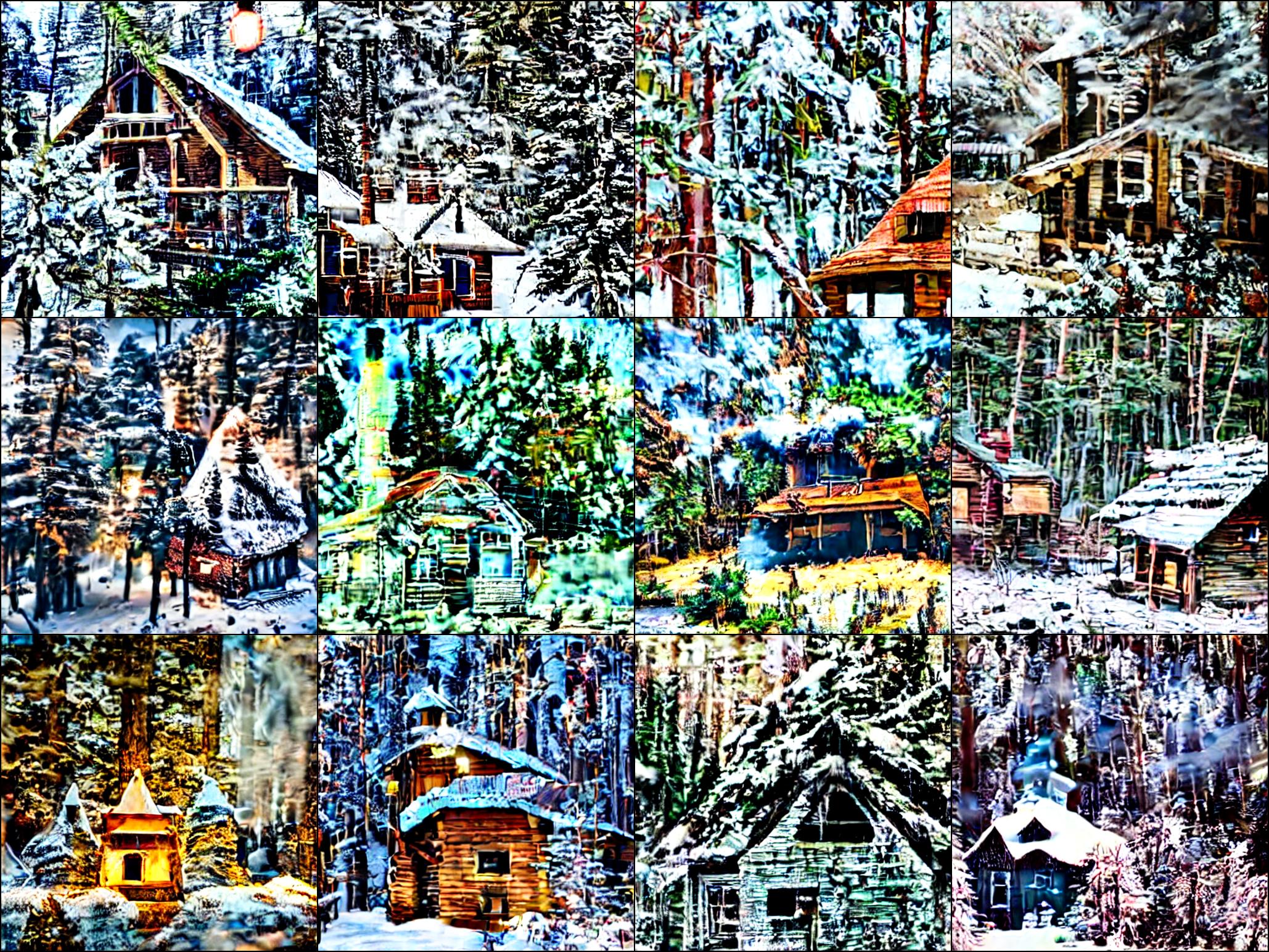}
        \label{}
    }
    \subfloat[][MPQ-DM]{
        \includegraphics[width=0.5\textwidth]{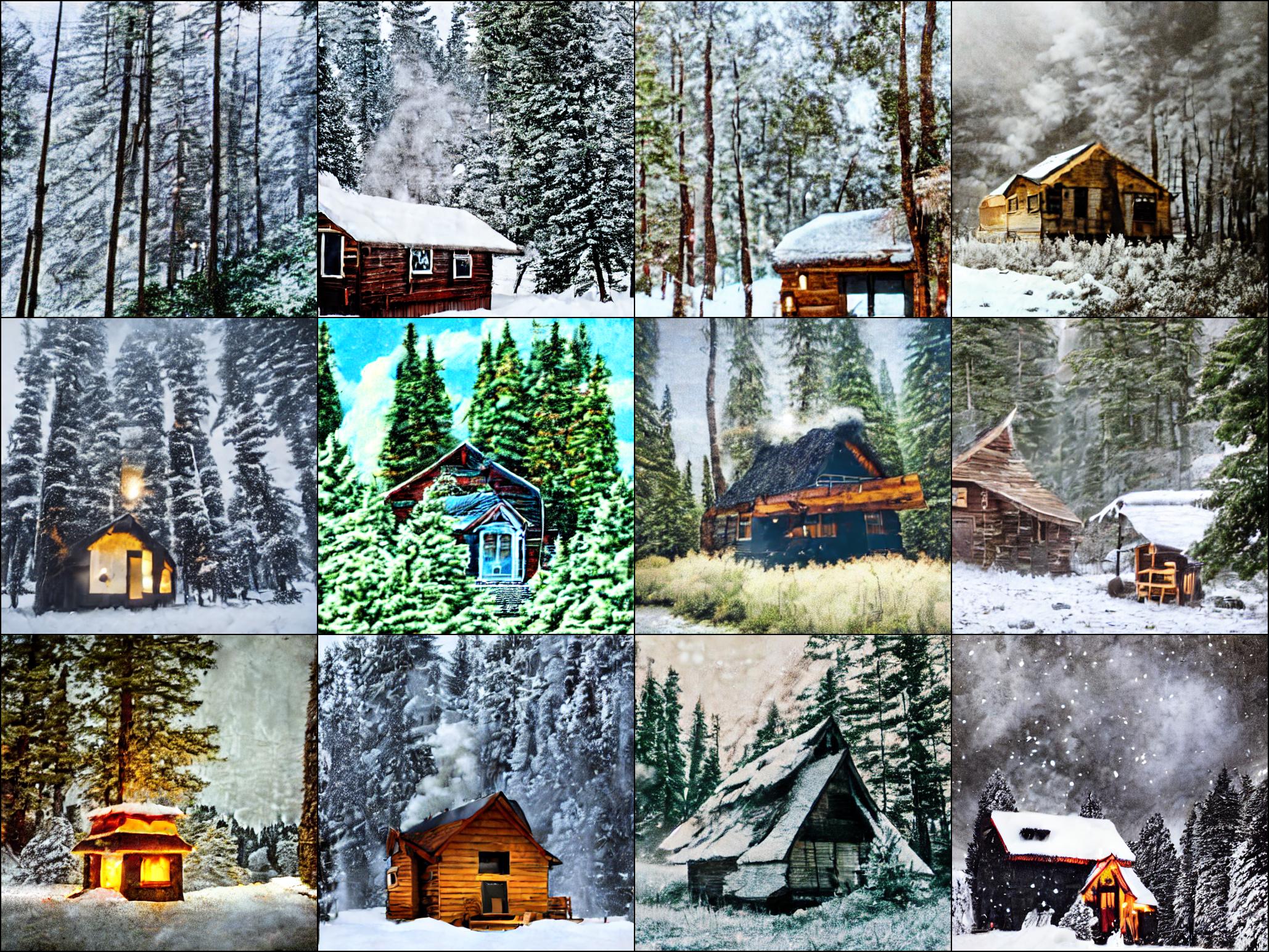}
        \label{}
    }
    \caption{Randomly generated samples W3A6 Stable Diffusion model with prompt “A cozy cabin nestled in a snowy forest with smoke rising from the chimney”.}
\end{minipage}
\end{figure*}

\begin{figure*}
\begin{minipage}{\textwidth}
    \centering
    \subfloat[][Baseline]{
        \includegraphics[width=0.5\textwidth]{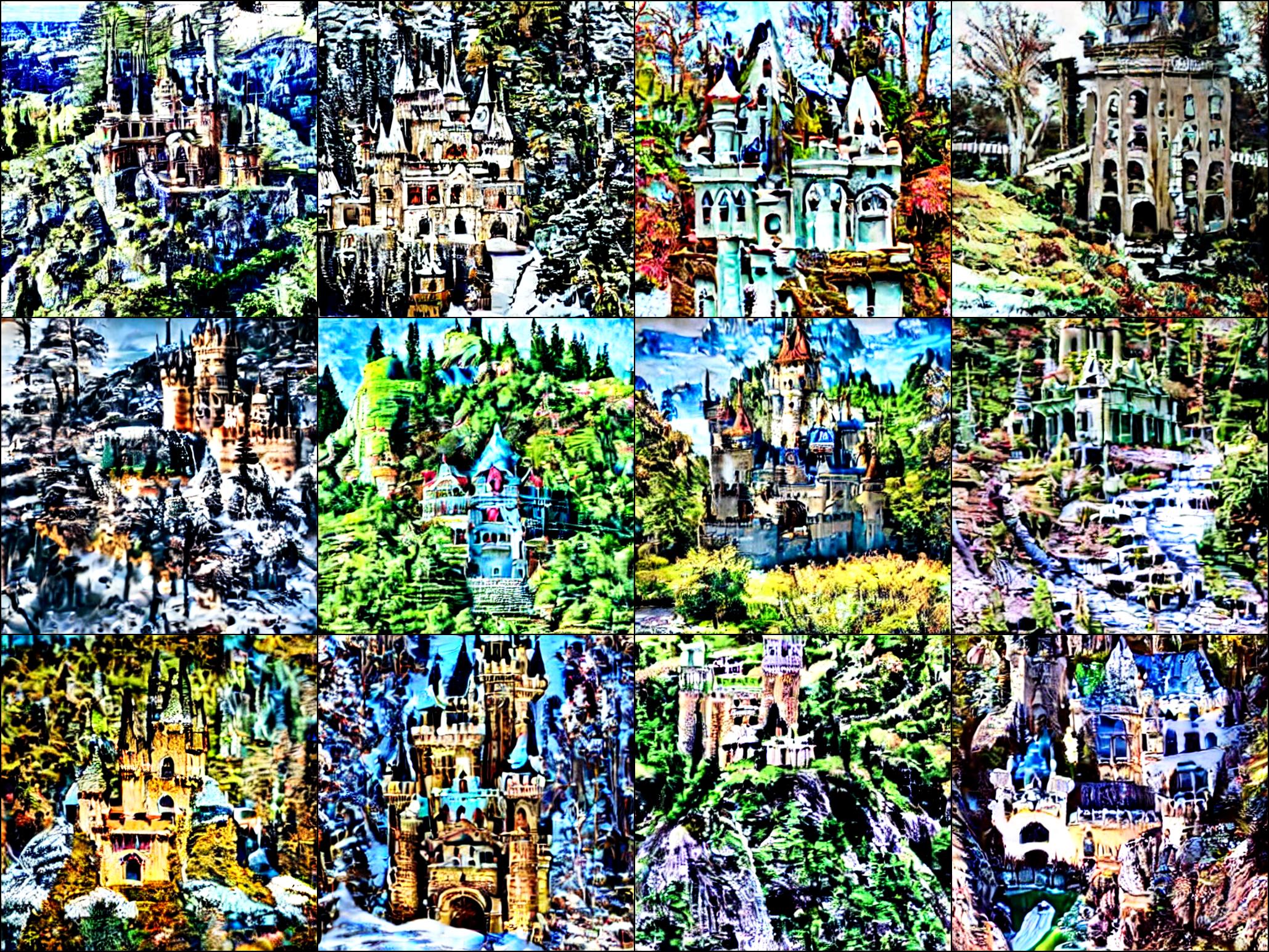}
        \label{}
    }
    \subfloat[][MPQ-DM]{
        \includegraphics[width=0.5\textwidth]{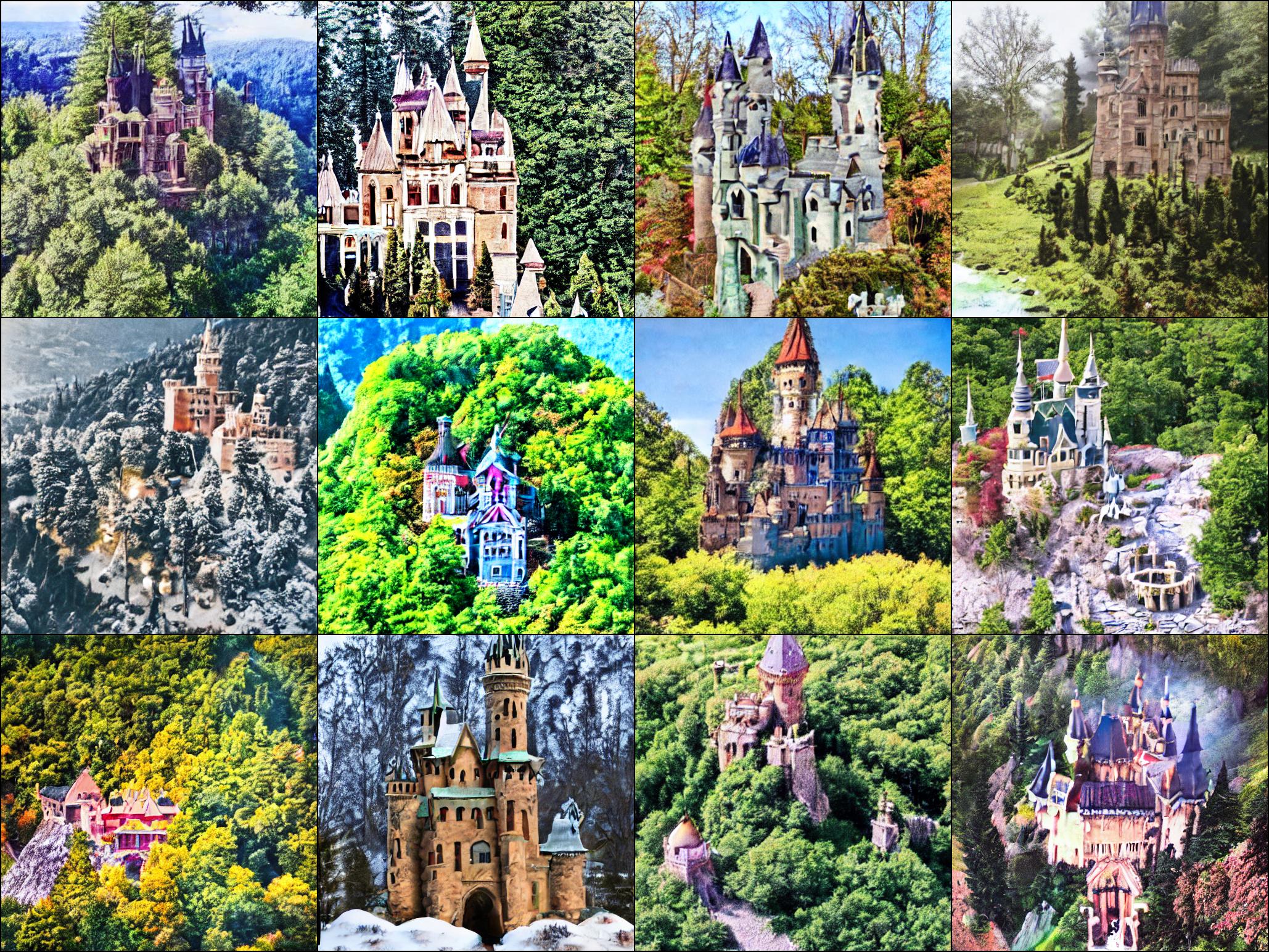}
        \label{}
    }
    \caption{Randomly generated samples W3A6 Stable Diffusion model with prompt “a
magical fairy tale castle on a hilltop surrounded by a mystical forest”.}
\end{minipage}
\end{figure*}

\begin{figure*}
\begin{minipage}{\textwidth}
    \centering
    \subfloat[][Baseline]{
        \includegraphics[width=0.5\textwidth]{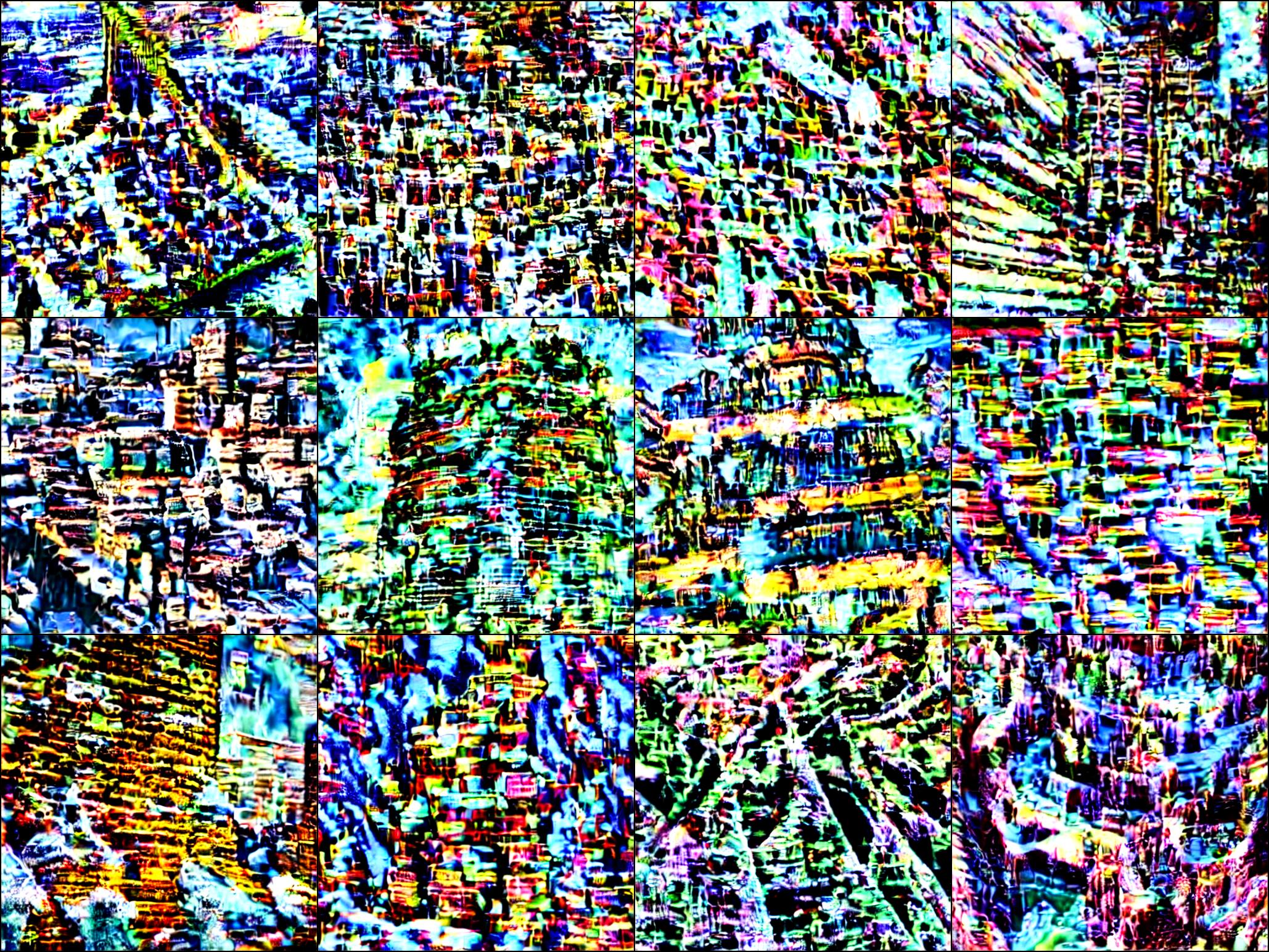}
        \label{}
    }
    \subfloat[][MPQ-DM]{
        \includegraphics[width=0.5\textwidth]{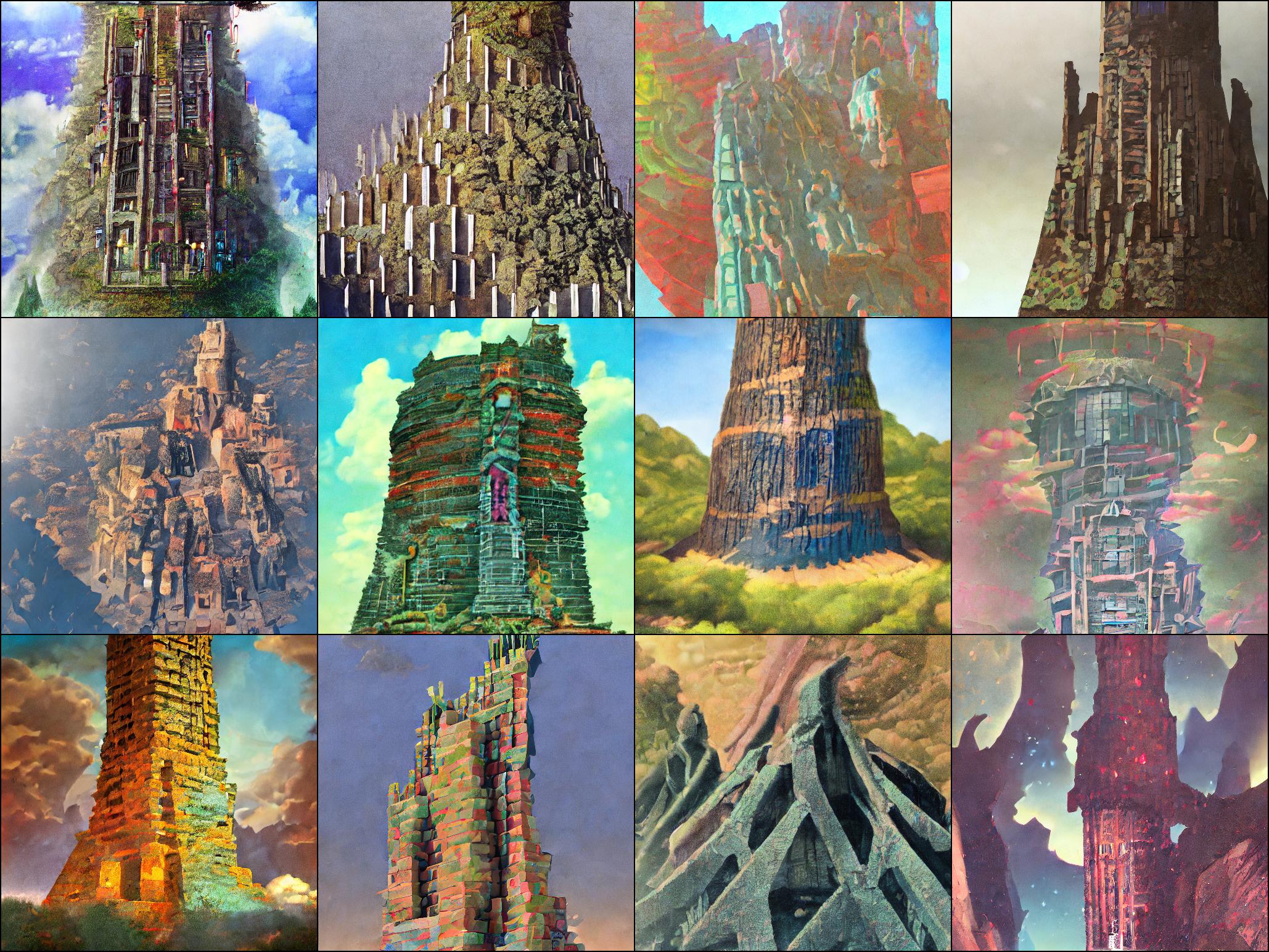}
        \label{}
    }
    \caption{Randomly generated samples W3A6 Stable Diffusion model with prompt “A digital illustration of the Babel tower, detailed, trending in artstation, fantasy vivid colors”.}
\end{minipage}
\end{figure*}

\begin{figure*}
\begin{minipage}{\textwidth}
    \centering
    \subfloat[][Baseline]{
        \includegraphics[width=0.5\textwidth]{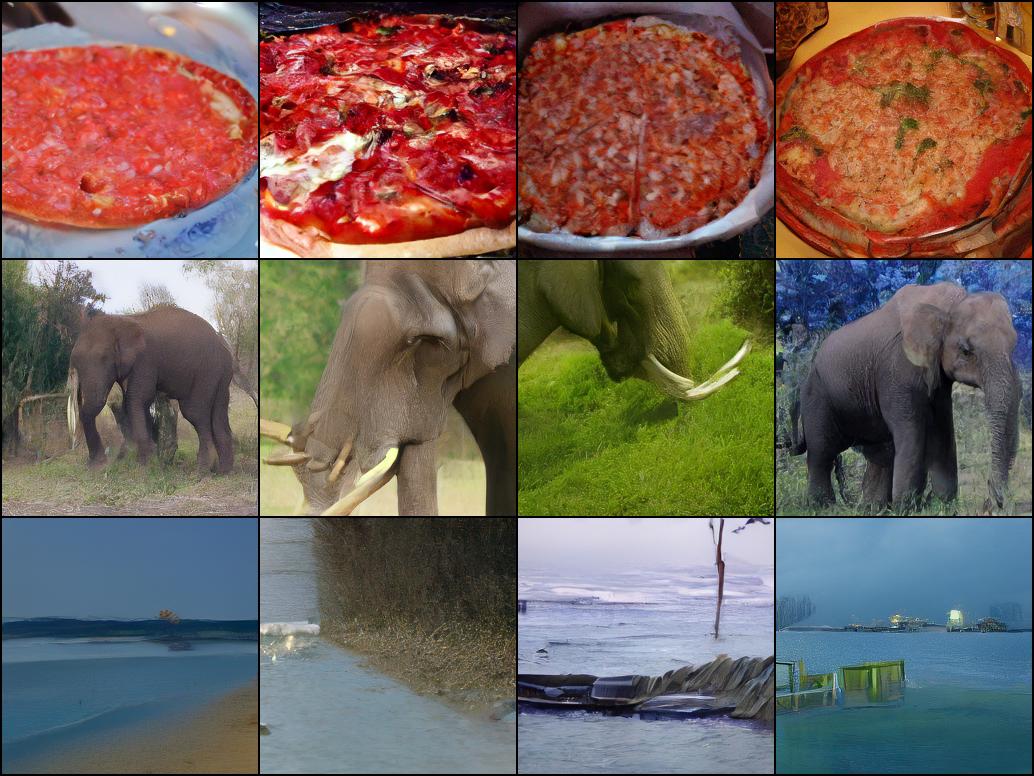}
        \label{}
    }
    \subfloat[][MPQ-DM]{
        \includegraphics[width=0.5\textwidth]{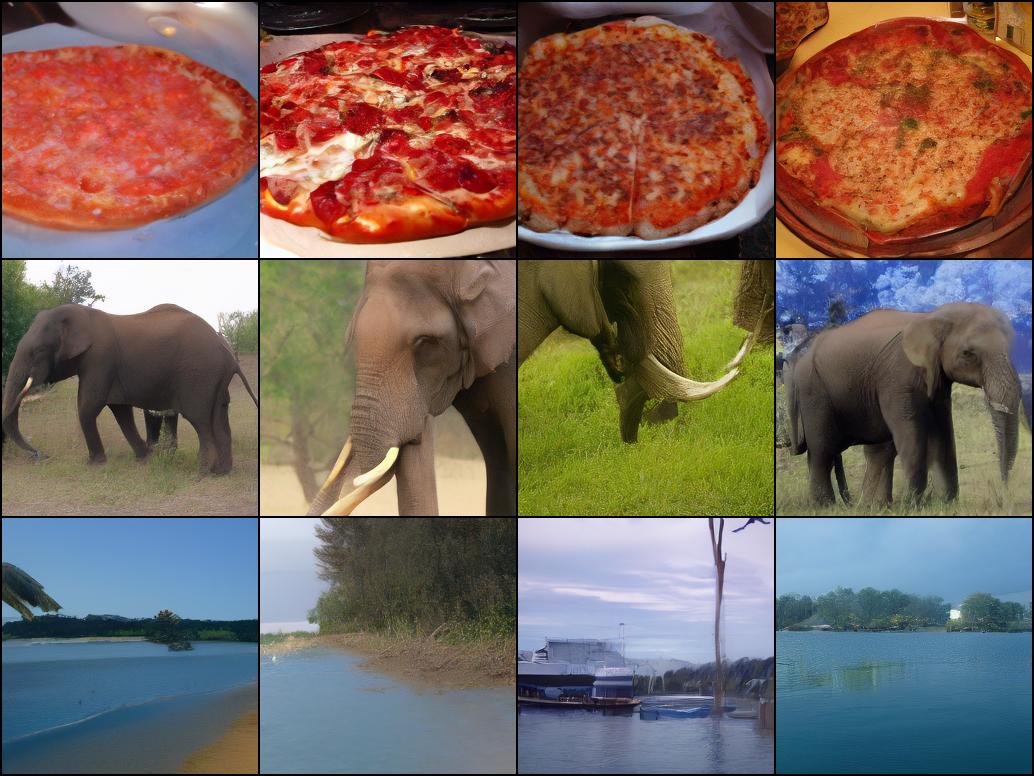}
        \label{}
    }
    \caption{Visualization of samples generated by W3A4 LDM model on ImageNet 256$\times$256.}
    \label{fig:imagenet_w3a4}
\end{minipage}
\end{figure*}

\begin{figure*}
\begin{minipage}{\textwidth}
    \centering
    \subfloat[][Baseline]{
        \includegraphics[width=0.5\textwidth]{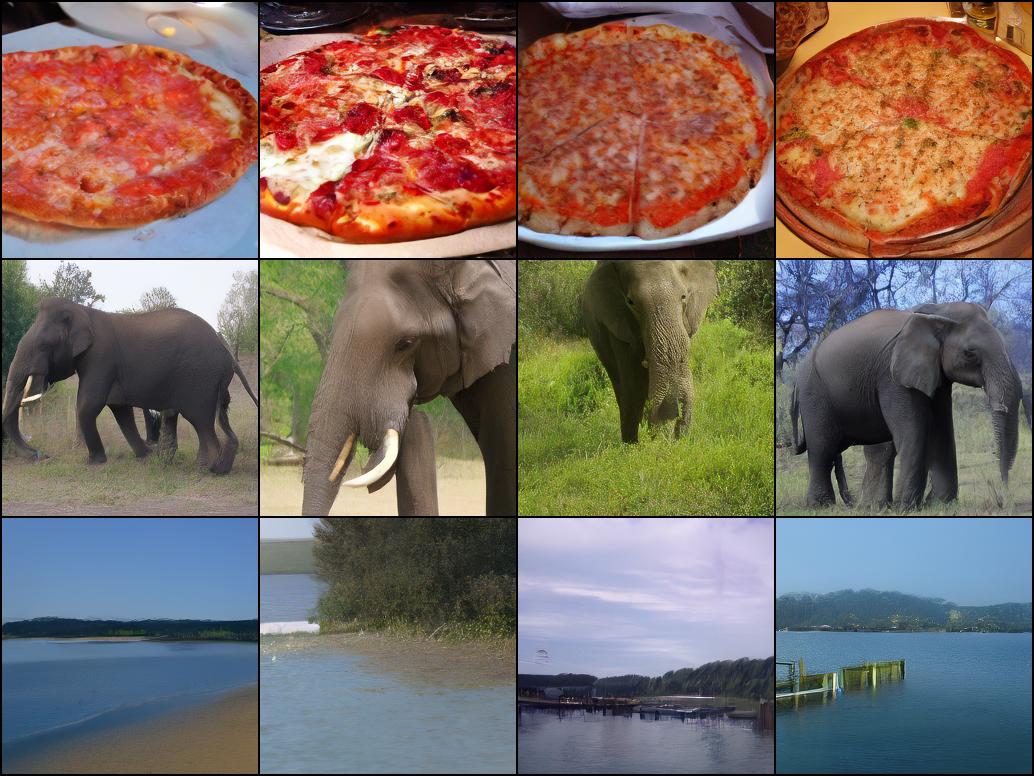}
        \label{}
    }
    \subfloat[][MPQ-DM]{
        \includegraphics[width=0.5\textwidth]{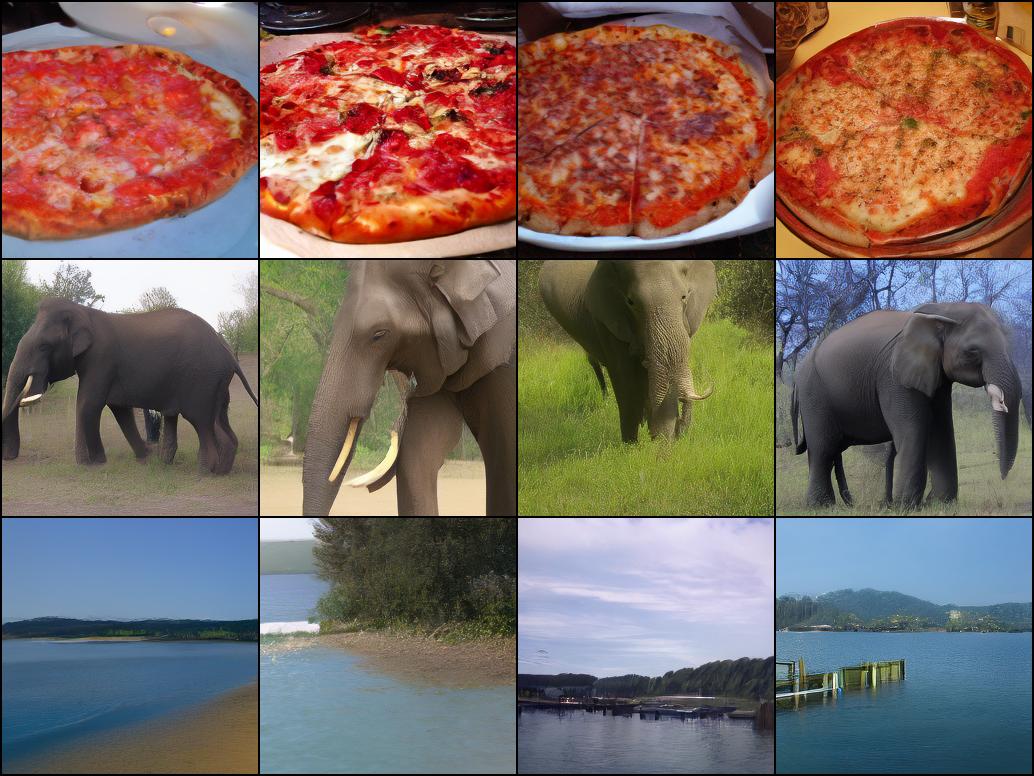}
        \label{}
    }
    \caption{Visualization of samples generated by W3A6 LDM model on ImageNet 256$\times$256.}
    \label{fig:imagenet_w3a6}
\end{minipage}
\end{figure*}

\begin{figure*}
\begin{minipage}{\textwidth}
    \centering
    \subfloat[][Baseline]{
        \includegraphics[width=0.5\textwidth]{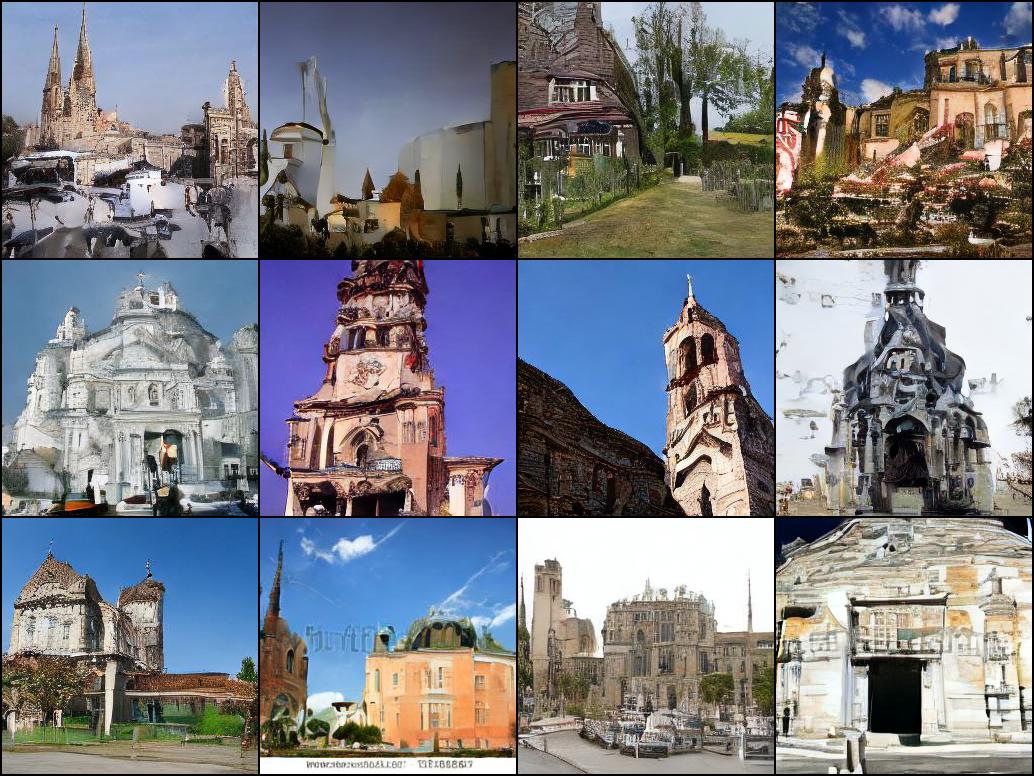}
        \label{}
    }
    \subfloat[][MPQ-DM]{
        \includegraphics[width=0.5\textwidth]{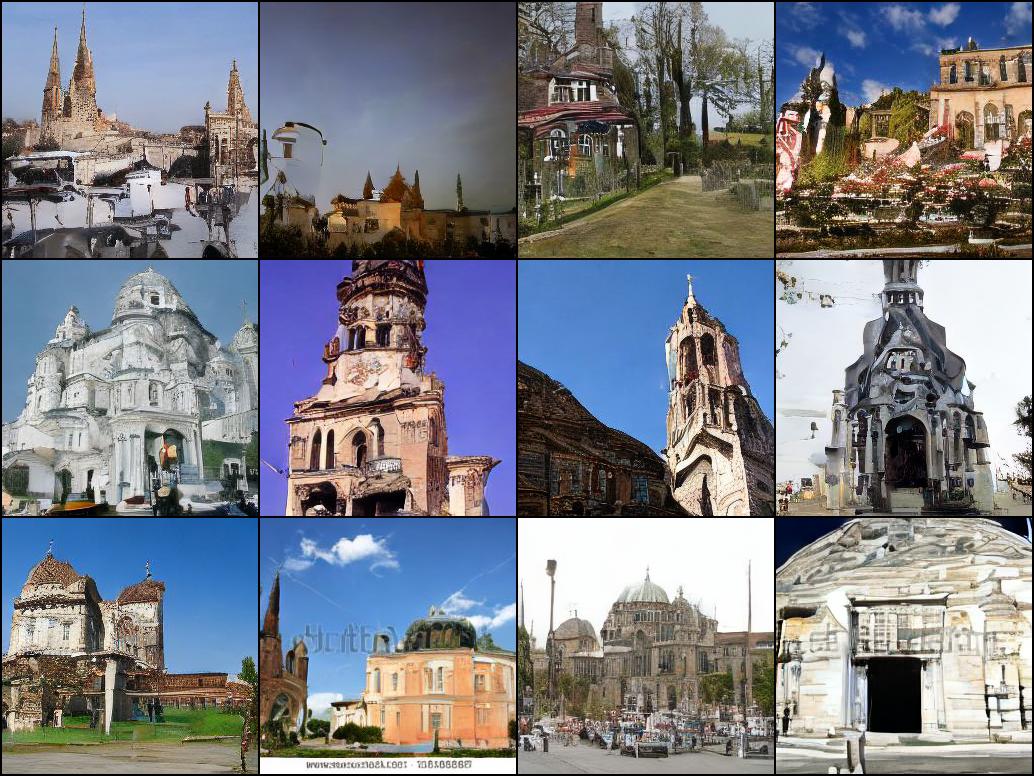}
        \label{}
    }
    \caption{Visualization of samples generated by W2A4 LDM model on LSUN-Churchs 256$\times$256.}
\end{minipage}
\end{figure*}

\begin{figure*}
\begin{minipage}{\textwidth}
    \centering
    \subfloat[][Baseline]{
        \includegraphics[width=0.5\textwidth]{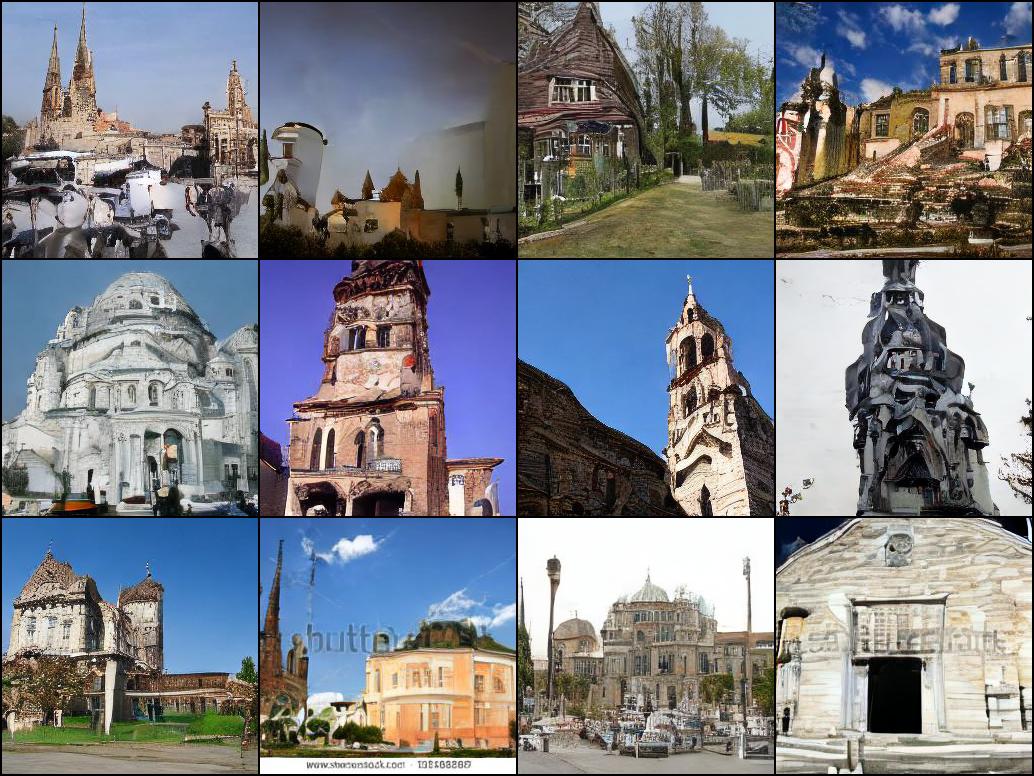}
        \label{}
    }
    \subfloat[][MPQ-DM]{
        \includegraphics[width=0.5\textwidth]{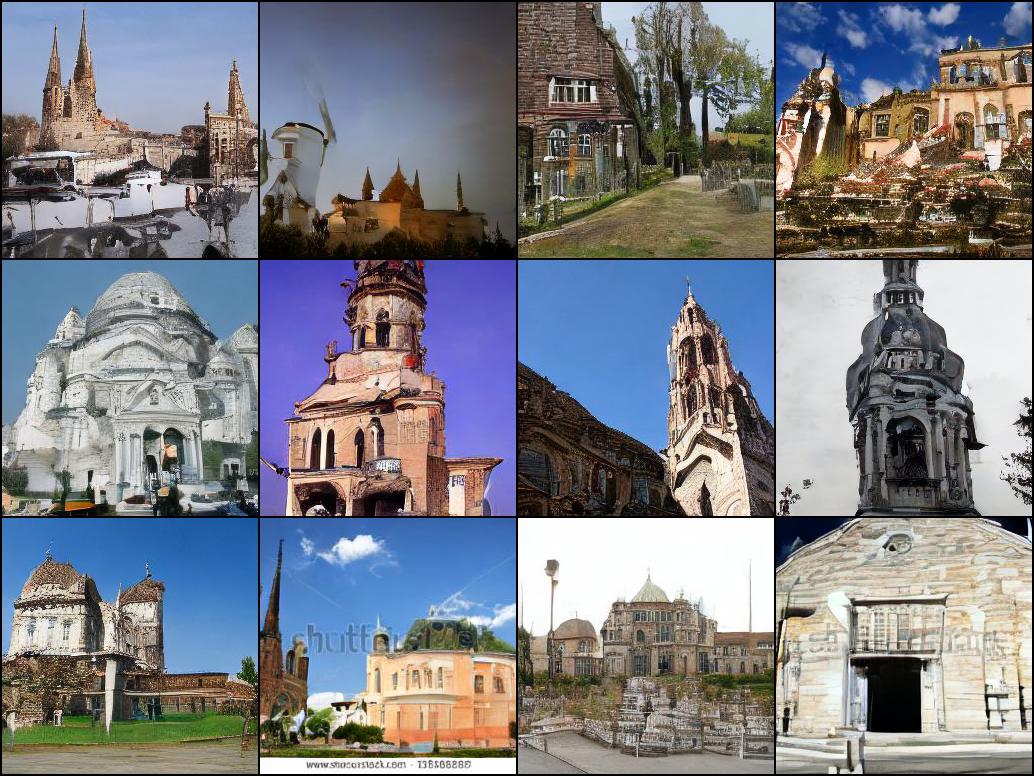}
        \label{}
    }
    \caption{Visualization of samples generated by W2A6 LDM model on LSUN-Churchs 256$\times$256.}
\end{minipage}
\end{figure*}

\begin{figure*}
\begin{minipage}{\textwidth}
    \centering
    \subfloat[][Baseline]{
        \includegraphics[width=0.5\textwidth]{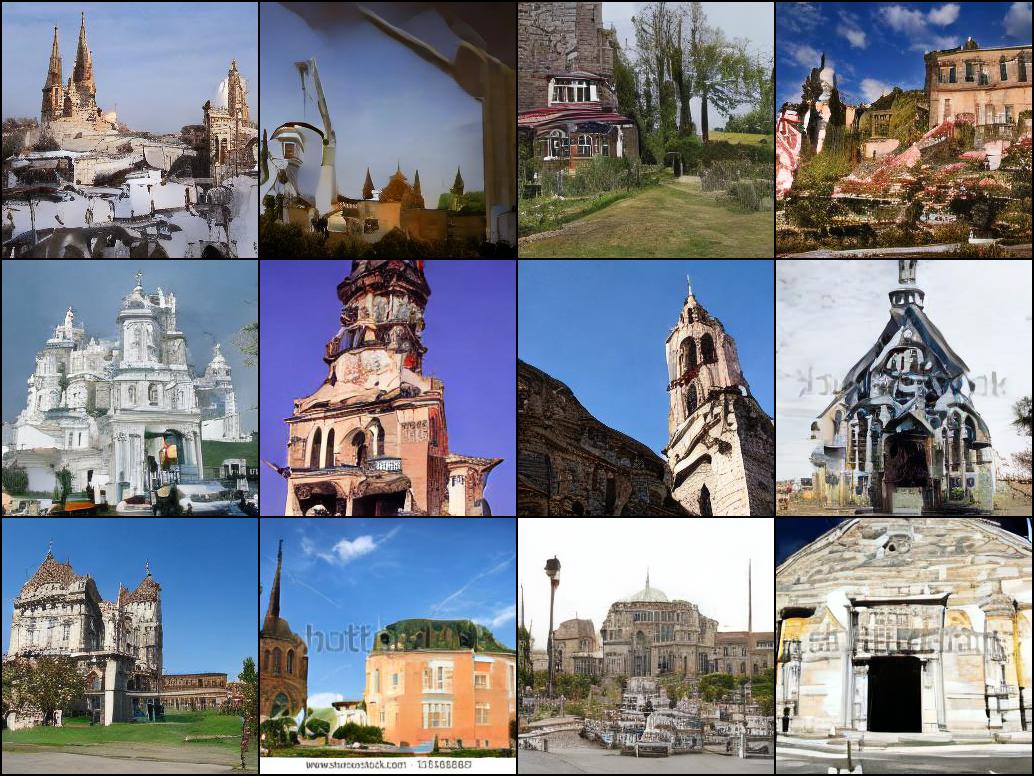}
        \label{}
    }
    \subfloat[][MPQ-DM]{
        \includegraphics[width=0.5\textwidth]{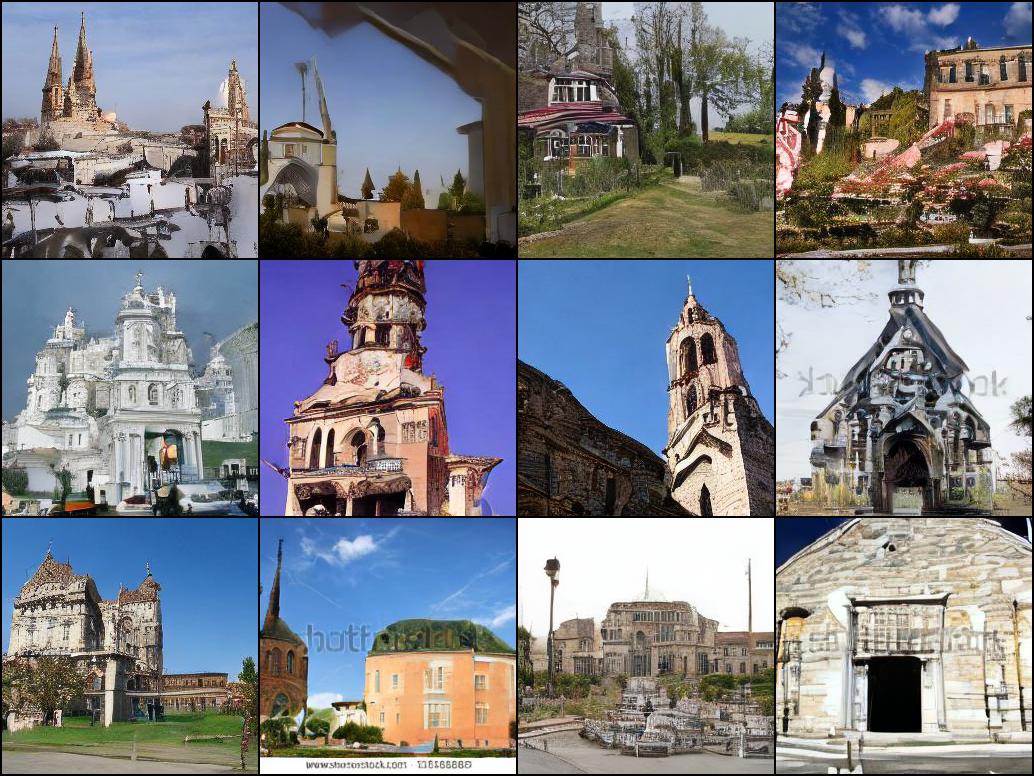}
        \label{}
    }
    \caption{Visualization of samples generated by W3A4 LDM model on LSUN-Churchs 256$\times$256.}
\end{minipage}
\end{figure*}

\begin{figure*}
\begin{minipage}{\textwidth}
    \centering
    \subfloat[][Baseline]{
        \includegraphics[width=0.5\textwidth]{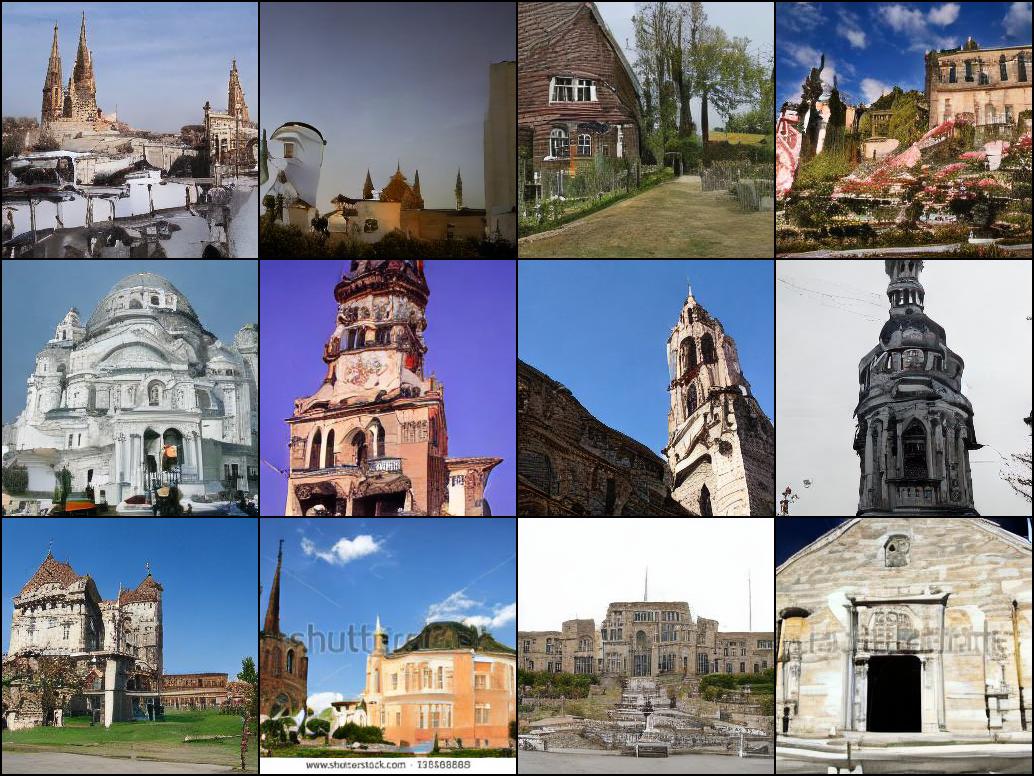}
        \label{}
    }
    \subfloat[][MPQ-DM]{
        \includegraphics[width=0.5\textwidth]{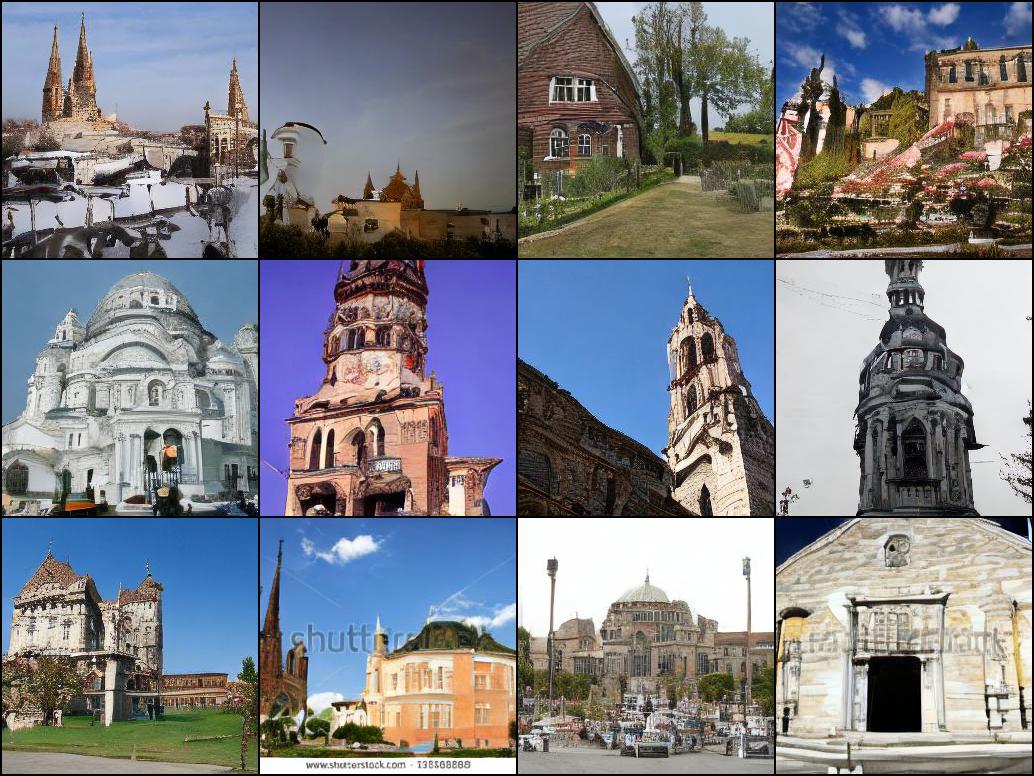}
        \label{}
    }
    \caption{Visualization of samples generated by W3A6 LDM model on LSUN-Churchs 256$\times$256.}
\end{minipage}
\end{figure*}

\begin{figure*}
\begin{minipage}{\textwidth}
    \centering
    \subfloat[][Baseline]{
        \includegraphics[width=0.5\textwidth]{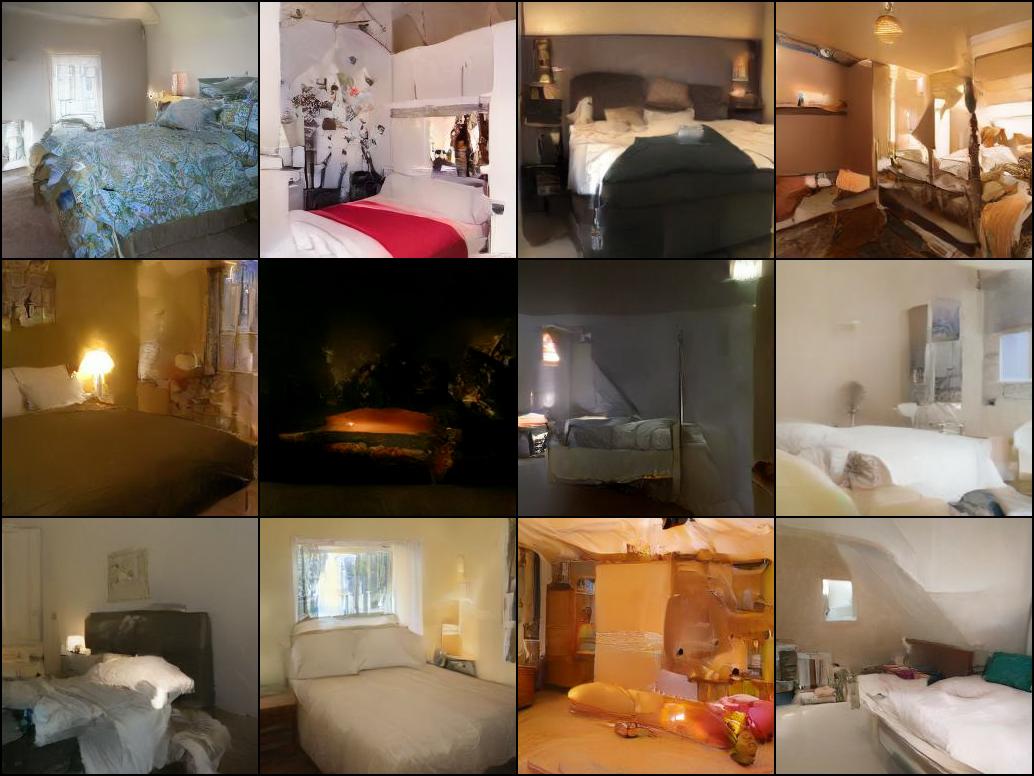}
        \label{}
    }
    \subfloat[][MPQ-DM]{
        \includegraphics[width=0.5\textwidth]{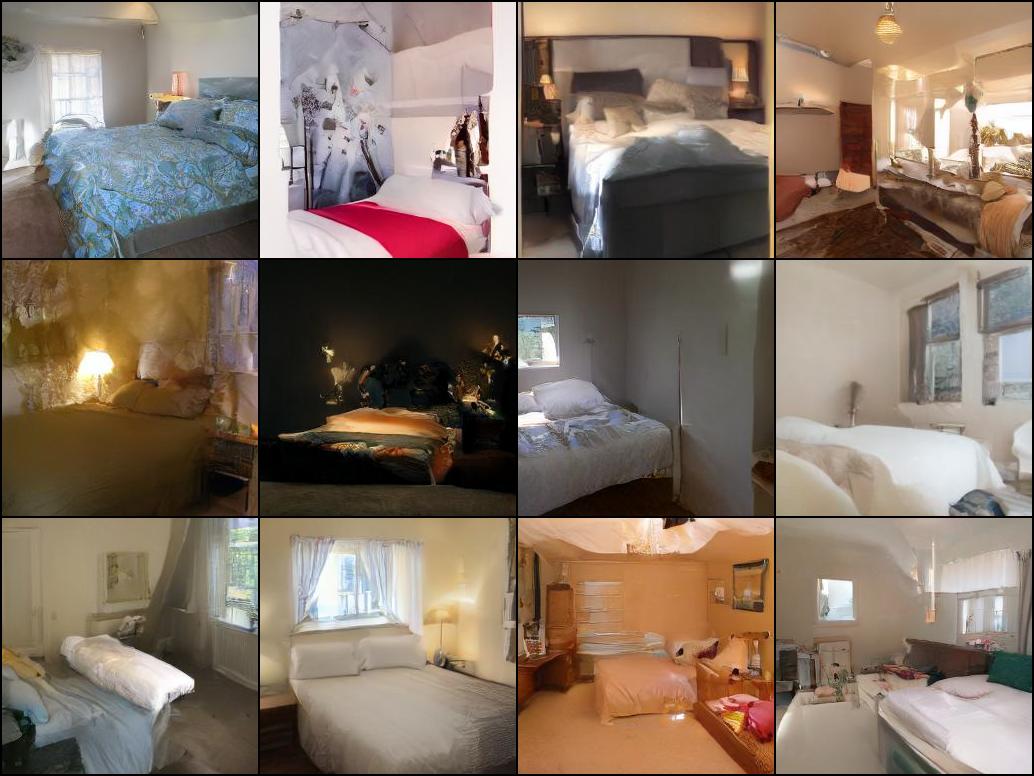}
        \label{}
    }
    \caption{Visualization of samples generated by W2A4 LDM model on LSUN-Bedrooms 256$\times$256.}
\end{minipage}
\end{figure*}

\begin{figure*}
\begin{minipage}{\textwidth}
    \centering
    \subfloat[][Baseline]{
        \includegraphics[width=0.5\textwidth]{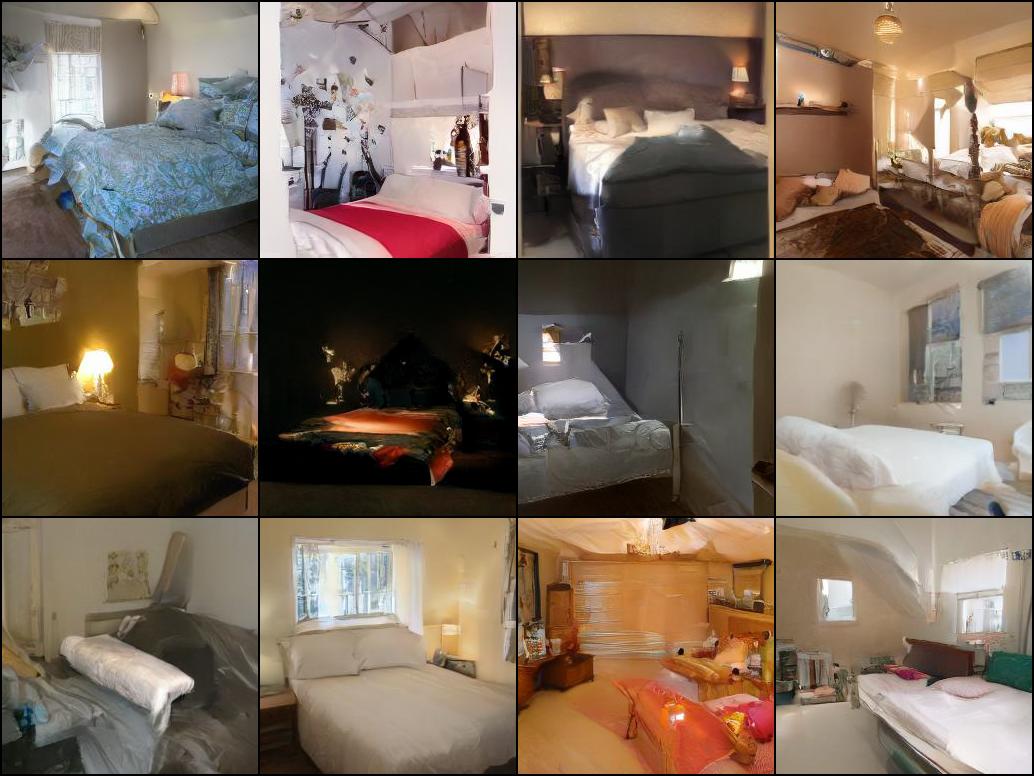}
        \label{}
    }
    \subfloat[][MPQ-DM]{
        \includegraphics[width=0.5\textwidth]{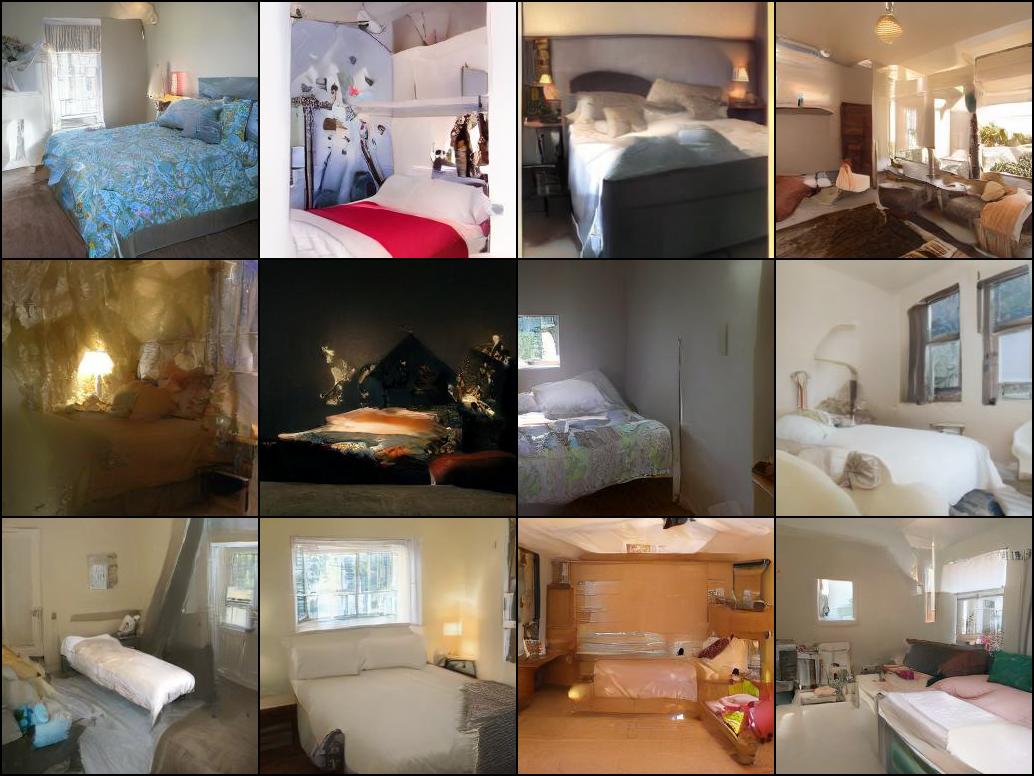}
        \label{}
    }
    \caption{Visualization of samples generated by W2A6 LDM model on LSUN-Bedrooms 256$\times$256.}
\end{minipage}
\end{figure*}

\begin{figure*}
\begin{minipage}{\textwidth}
    \centering
    \subfloat[][Baseline]{
        \includegraphics[width=0.5\textwidth]{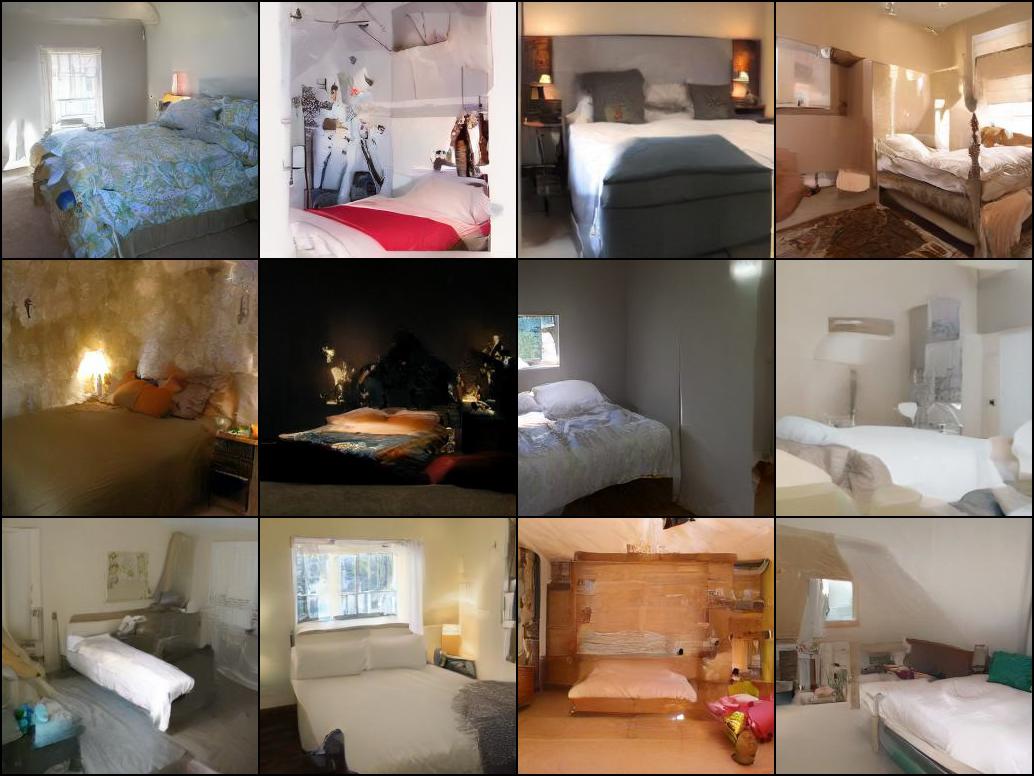}
        \label{}
    }
    \subfloat[][MPQ-DM]{
        \includegraphics[width=0.5\textwidth]{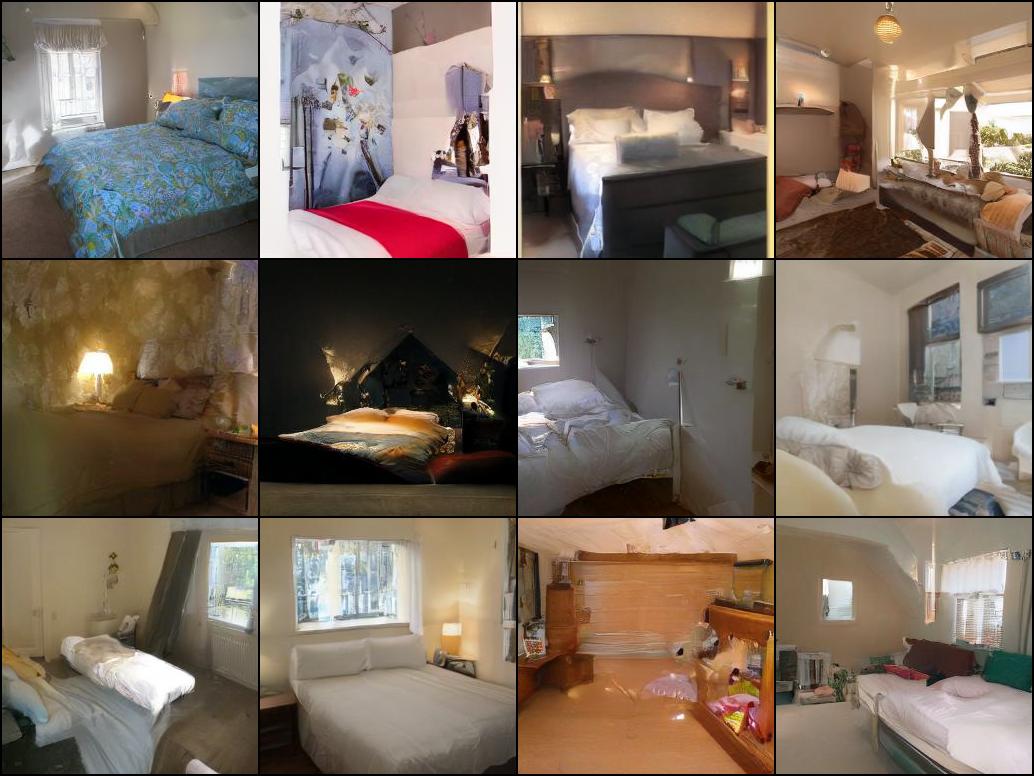}
        \label{}
    }
    \caption{Visualization of samples generated by W3A4 LDM model on LSUN-Bedrooms 256$\times$256.}
\end{minipage}
\end{figure*}

\begin{figure*}
\begin{minipage}{\textwidth}
    \centering
    \subfloat[][Baseline]{
        \includegraphics[width=0.5\textwidth]{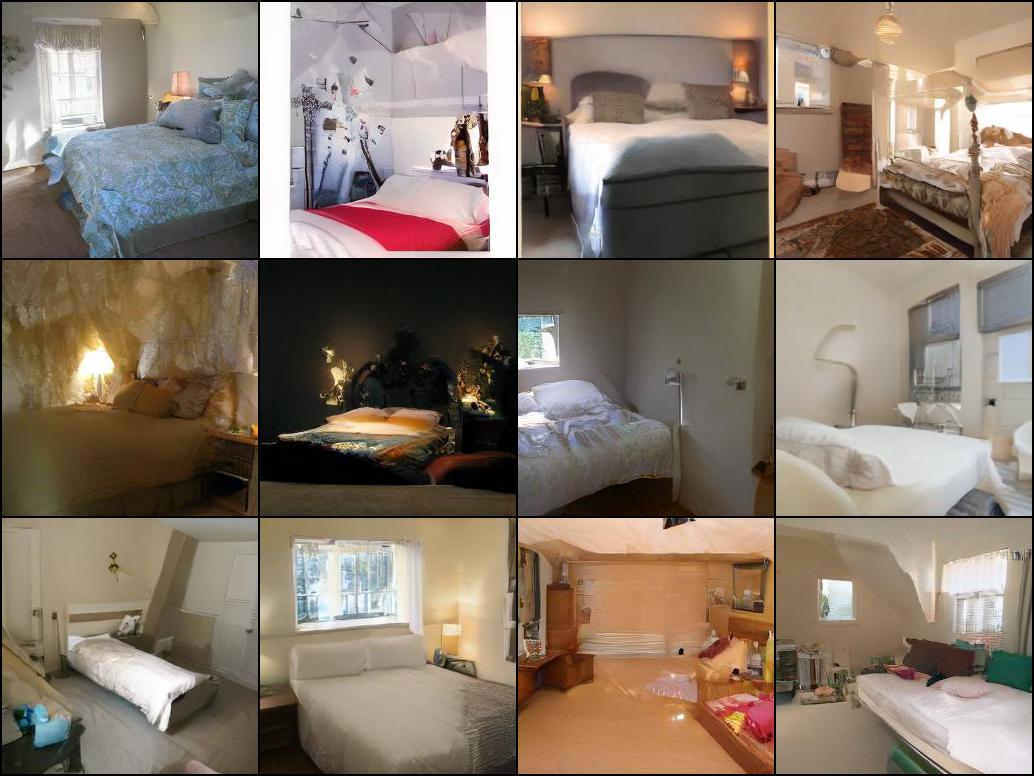}
        \label{}
    }
    \subfloat[][MPQ-DM]{
        \includegraphics[width=0.5\textwidth]{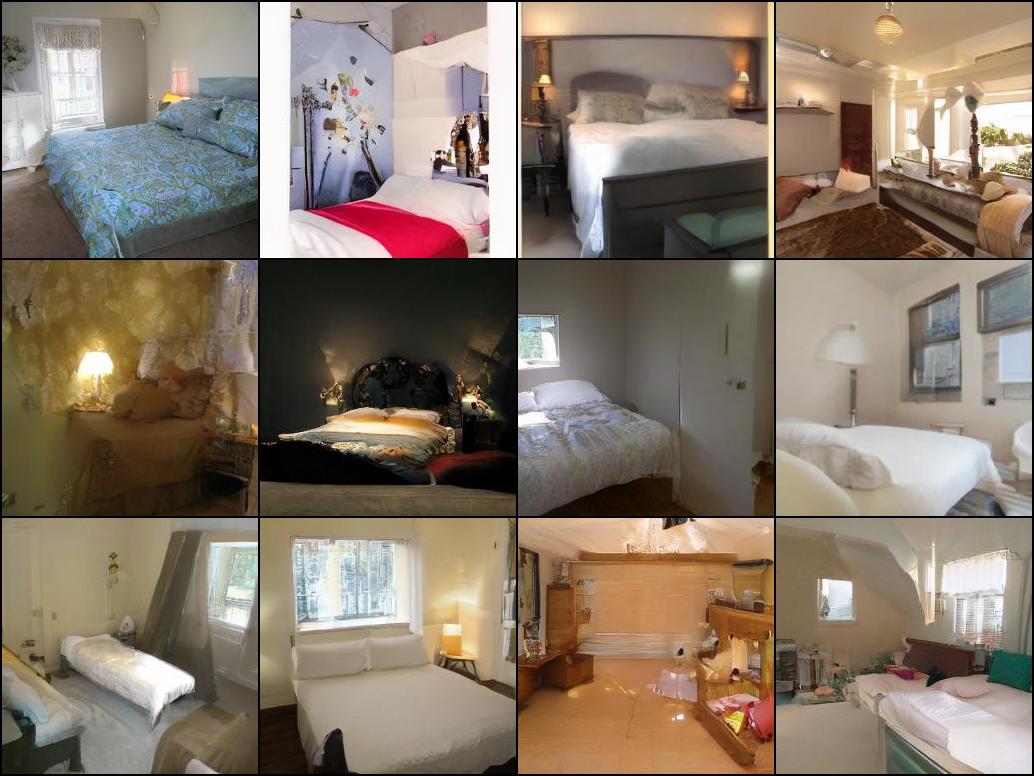}
        \label{}
    }
    \caption{Visualization of samples generated by W3A6 LDM model on LSUN-Bedrooms 256$\times$256.}
\end{minipage}
\end{figure*}